\documentclass{article}
\usepackage{ijcai21}

\usepackage{times}

\usepackage{soul}
\usepackage{url}
\usepackage[hidelinks]{hyperref}
\usepackage[utf8]{inputenc}
\usepackage[small]{caption}
\usepackage{graphicx}
\usepackage{amsmath}
\usepackage{booktabs}
\usepackage{natbib}
\usepackage{algorithm}
\usepackage{algorithmic}
\DeclareMathOperator*{\argmin}{arg\,min}

\usepackage{ntheorem}
\theoremseparator{:}
\newtheorem{hyp}{Hypothesis}

\newtheorem{fac}{Fact}

\title{Knowledge Capture and Replay for Continual Learning}

\author {
        Saisubramaniam Gopalakrishnan \textsuperscript{\rm 1},
       Pranshu Ranjan Singh \textsuperscript{\rm 1},
       Haytham Fayek \textsuperscript{\rm 3},
       Savitha Ramasamy \textsuperscript{\rm 1, 2},
        Arulmurugan Ambikapathi \textsuperscript{\rm 1, 2} \\

\affiliations 
    \textsuperscript{\rm 1} Institute for Infocomm Research (I$^2$R), A*STAR \\
    \textsuperscript{\rm 2} Artificial Intelligence, Analytics And Informatics (AI$^3$), A*STAR \\
    \textsuperscript{\rm 3} RMIT University \\
    \{g\_saisubramaniam, pranshurs, ramasamysa, arul\}@i2r.a-star.edu.sg, haytham.fayek@ieee.org
}

\begin{document}
\maketitle
\begin{abstract}
Deep neural networks have shown promise in several domains, and the learned data (task) specific information is implicitly stored in the network parameters. Extraction and utilization of encoded knowledge representations is vital when data is no longer available in the future, especially in a continual learning scenario. In this work, we introduce {\em flashcards}, which are visual representations that {\em capture} the encoded knowledge of a network as a recursive function of predefined random image patterns. In a continual learning scenario, flashcards help to prevent catastrophic forgetting and consolidating knowledge of all the previous tasks. Flashcards need to be constructed only before learning the subsequent task, and hence, independent of the number of tasks trained before. We demonstrate the efficacy of flashcards in capturing learnt knowledge representation (as an alternative to original dataset), and empirically validate on a variety of continual learning tasks: reconstruction, denoising, task-incremental learning, and new-instance learning classification, using several heterogeneous benchmark datasets. Experimental evidence indicates that: (i) flashcards as a replay strategy is { \em task agnostic}, (ii) performs better than generative replay, and (iii) is on par with episodic replay without additional memory overhead.
\end{abstract}

% \begin{figure}[htp]
% 	\centering
% 	\includegraphics[width=0.48\textwidth]{figure/main_architecture_fig1.png}
% 	\caption{Flashcards demonstration. Step (a): Trained autoencoder (AE) for Task 1 (N-1); Step (b): Construct flashcards from frozen AE using maze patterns via iterative passes; Step (c): Replay using flashcards on new network to remember Task N-1; Step (d) Replay using flashcards of past while training for new Task N.}
% 	\label{proposed_framework_main}
% 	\vspace{-0.3cm}
% \end{figure}
% \vspace{-0.2cm}

\begin{figure}[t]
	\centering
	\includegraphics[width=0.48\textwidth]{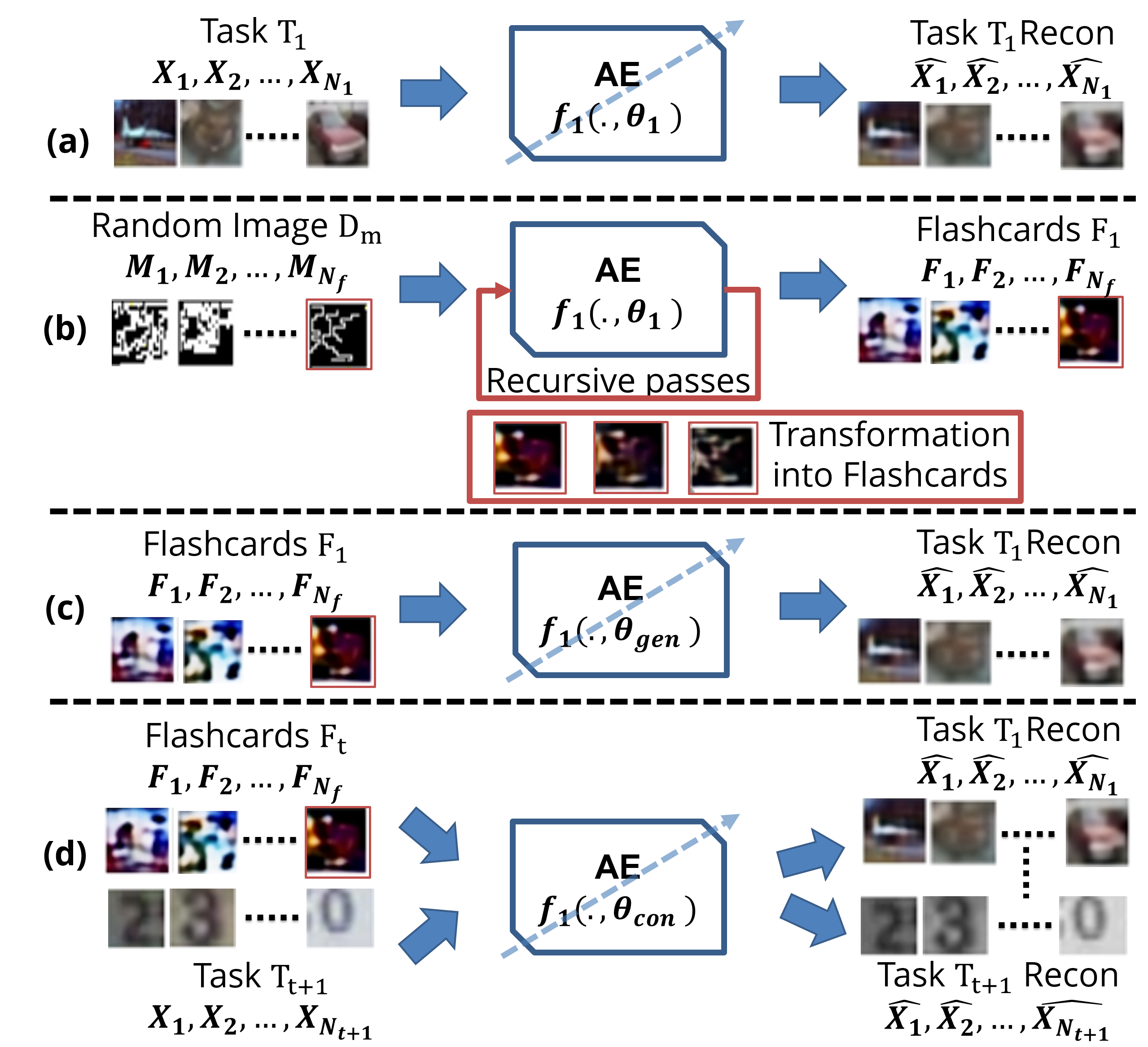}
	\caption{Flashcards for knowledge capture and replay. Step (a): Trained autoencoder (AE) for Task $T_1$; Step (b): Construct flashcards from frozen AE using maze patterns via recursive passes; Step (c): Replay using flashcards on a new network to remember Task $T_1$; Step (d) Replay using flashcards $F_{t}$ while training for Task $T_{t+1}$.}
	\label{proposed_framework_main}
% 	\vspace{-0.1cm}
\end{figure}
% \vspace{-0.3cm}

\section{Introduction}
Deep neural networks are successful in several applications \citep{CV2012,NLP2019,Pattrec2014}, however, it remains a challenge to extract and reuse the knowledge embedded in the representations of the trained models, in a benign manner, for similar or other downstream scenarios \citep{liubing2020}. Approaches such as transfer learning \citep{zhuang2019comprehensive} and knowledge distillation \citep{KDsurvey_2020} enable representational or functional translation \citep{KD2015} from one model to the other. Importantly, knowledge captured from the past should be preserved in some representation with low computational and memory expense, especially, when the past data is no longer available.
% \citep{braininspired2020}.

% In practice, 
Learning new tasks affects retention of previous knowledge \citep{robins1995catastrophic}. Continual learning (CL) \citep{thrun1996learning} was proposed as a candidate for such retention, and is gaining attention recently \citep{clreview}. Essentially, CL aims to make deep neural networks learn continually from a sequence of tasks efficiently, without catastrophically forgetting the past sequences/tasks. There is a need to develop a reliable method that can effectively capture and re-purpose knowledge from a learned model, which can then be exploited in the CL setting. Also, while CL approaches are predominantly studied on classification (class/task incremental learning), its potential remains relatively less explored in other scenarios such as task agnostic unsupervised reconstruction, denoising, new instance learning, etc.
% , where deep neural networks are quite successful. 

% , to preserve and reuse that knowledge from past tasks while learning the future tasks. 

\begin{figure*}[t]
	\centering
	\includegraphics[width=1\textwidth]{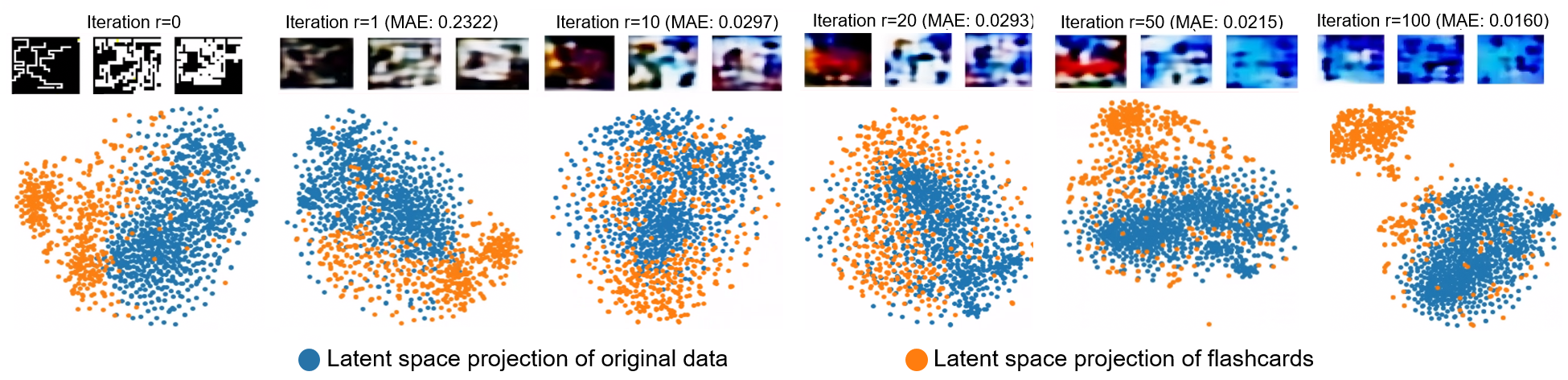}
	\vspace{-0.25cm}
	\caption{A sample of three Flashcards ($F_t$) constructions over recursive iterations, with Cifar10 as an example, difference in MAE between successive iterations with iteration number and t-SNE of latent space in 2D. It can be observed that the initial (raw) input maze patterns adapt to the texture captured by the model, as the number of recursive iterations increase (Iteration $r=10,20$). Also, the corresponding latent space clusters get closer. However, on further repeated passes ($r\geq50$), the reconstruction deteriorates / smooth out (as MAE keeps reducing) and the two clusters drift apart again (P1 is satisfied but P2 is not). Best viewed in color.}
	\label{tsne_main}
	\vspace{-0.4cm}
\end{figure*}

In this work, we introduce \textit{flashcards} for capturing consolidated learned knowledge representations of all the previous tasks. Random image patterns, when passed recursively through a trained autoencoder, capture its representations with each pass, and are transformed into flashcards.
%When random image patterns are passed recursively through a trained autoencoder, representations of the data learnt by the network parameters gets captured with each pass, and transformed into flashcards. 
A schematic illustration is shown in Figure~\ref{proposed_framework_main}. We evaluate flashcards' potential for capturing representations of the data (as seen by the network) and show their usefulness for replay while learning continually from several heterogeneous datasets, for a variety of downstream applications such as reconstruction, denoising, task-incremental, and new-instance classification.

% In this work, we introduce \textit{flashcards} to capture learned knowledge representations in a neural network which can then potentially be used to transfer knowledge across tasks. Flashcards are opaque random image patterns that capture the representations of a neural network by recursively passing through a trained autoencoder model. A schematic illustration for flashcard generation and its application in CL scenario is shown in Figure~\ref{proposed_framework_main}. We first demonstrate the ability of flashcards in efficiently capturing representations from a trained model and evaluate their captured knowledge representations. Next, we show their usefulness for replay in a CL setting, while learning continually from sequentially streaming heterogeneous datasets, namely, MNIST, Fashion MNIST, Cifar10, SVHN and Omniglot. While the vast majority of CL algorithms are mostly suited for classification tasks, we anchor our demonstrations on unsupervised representations. Our evaluation studies emphasize the usefulness of flashcards in a CL setting by using flashcards as replay mechanisms, against prior state-of-the-art replay strategies. Importantly, we demonstrate the effectiveness of flashcards in capturing representations in CL setting for a variety of downstream applications such as reconstruction, denoising, task incremental learning and new instance learning.

\section{Knowledge Capture and Replay}% 
\label{sec_methods}
We first introduce the notion of knowledge capture using flashcards and describe their construction from a trained autoencoder (AE)\footnote{AE is used {\em only} to construct flashcards which can be used for different CL applications, including classification (see Section~\ref{perf}).}. Consider training an AE for reconstruction task $T_t$ using dataset $D_{t}$ containing $N_t$ samples, where $D_{t} = \left\{\mathbf{X}_{1},\mathbf{X}_{2},\ldots, \mathbf{X}_{N_t}\right\}\subset\Re^{k \times l \times c}$,
% Here, $\mathbf{X}_{i}\in \Re^{k \times l \times c}$ 
is the set of training image samples with $k$ rows, $l$ columns, and $c$ channels.
% and $c_i \in \{1,2,\ldots,s\}$ denotes the corresponding class labels among $s$ classes. In the unsupervised scenario considered here, the class labels are irrelevant. However, the labels in the dataset can be used to define sequential tasks (grouping of classes) within a dataset. 
For the task $T_t$, the AE is trained to maximize the likelihood $P(\mathbf{X}|{{{\boldsymbol \theta}_t}})$, $\forall \mathbf{X}\in{D_{t}}$ using the conventional 
% mean square error or 
mean absolute error (MAE; Eqn. \eqref{AE})  between the original and reconstructed images:
% as the loss function:
\begin{eqnarray}
\min_{{{\boldsymbol \theta}_t}} \frac{1}{N_t} \sum_{n=1}^{N_t} |\mathbf{X}_n- \widehat{\mathbf{X}}_n|, \label{AE}
\end{eqnarray}
% \begin{eqnarray}
% \min_{{{\boldsymbol \theta}_t}} \frac{1}{N_t} \sum_{n=1}^{N_t} L_{abs}(\mathbf{X}_n, \widehat{\mathbf{X}}_n), \label{AE}
% \end{eqnarray}
% \begin{eqnarray}
% L_{abs}(\mathbf{X}_n, \widehat{\mathbf{X}}_n)~=\frac{1}{m~n~c} \sum_{i=1}^{m}\sum_{j=1}^n\sum_{k=1}^c \Big{|}\mathbf{X}_n[i, j, k]- \nonumber \\
% \widehat{\mathbf{X}}_n[i, j, k]\Big{|}, \label{absolute-error}
% \end{eqnarray}
where ${{{\boldsymbol \theta}_t}}$ is the AE network parameters (weights, biases, and batch norm parameters) for task $T_t$, $\mathbf{X}_n$ is a sample from the empirical data distribution of ${P}(X)$,
% $\mathbf{X}_n \sim P(\mathbf{X})$ 
and $\widehat{\mathbf{X}}_n =f_{t}(\mathbf{X}_n,{{{\boldsymbol \theta}_t}})$ is the 
% are the true sample belonging to the distribution of $D_{t}$ and the 
reconstructed sample.
% respectively, 
$|\mathbf{X}_n- \widehat{\mathbf{X}}_n|$ denotes the pixel-wise mean absolute error between $\mathbf{X}_n$ and $\widehat{\mathbf{X}}_n$, and $f_{t}(\cdot)$ is the function approximated by the AE to learn (reconstruct) task $T_{t}$. In the above conventional learning setup, 
% it is obvious that 
the parameters $\mathbf{\theta}_t$ of an AE network aims to model the knowledge in the data $D_{t}$, such as {\em shape, texture, color} of the images. 
% Now, the following intriguing questions arise: Can this knowledge be suitably `captured’ as representations from the trained AE network? Can such knowledge representations \emph{alone} be then used to train a new network that has the performance close to the network trained with original data ($D_{t}$)? If such knowledge can be effectively captured from a trained network, then it can be used in multiple cases such as: (i) Training a different architecture (smaller or larger capacity) by just using the representations captured from the current network; (ii) In a CL scenario to remember information related to previous tasks, etc.
% The notion and procedure to construct flashcards is discussed next.
% In other words, for a sample $\mathbf{x}' \sim P(\mathbf{x})$ belonging to the distribution of $D_{T}$, the corresponding reconstructed output $\widehat{\mathbf{x}}'$ is:
% \begin{equation}
%     \widehat{\mathbf{x}}' = f_{t}(\mathbf{x}',{{{\boldsymbol \theta}_{t}}}).
% \end{equation}
% As before, $f_{t-1}(\cdot)$ is the function approximated by the AE to learn task $T_{t-1}$. Before, proceeding to learn the next task $D_t$, the weight activations of ${{{\boldsymbol \theta}_{t-1}}}$ need to be preserved and such knowledge should be transferred to the subsequent learning so as to avoid catastrophic forgetting of the previous task. 
Let the reconstruction error of the trained AE be bounded by [$\epsilon_1$,$\epsilon_2$], i.e.,
\begin{equation}
  \epsilon_1 \leq | \mathbf{X}_n -\widehat{\mathbf{X}}_n | \leq \epsilon_2,~\forall  \mathbf{X}_n \in D_{t} \sim P(\mathbf{X}) \label{AE_recon}
\end{equation}

Let $P(\mathbf{M})$ be a different, but well-defined distribution in the same dimensional space as $P(\mathbf{X})$ (i.e., $ \Re^{k \times l \times c}$), and $D_{m} = \left\{\mathbf{M}_{1},\mathbf{M}_{2},\ldots\mathbf{M}_{N_f}\right\}$, where $\mathbf{M}_{i} \in \Re^{k \times l \times c}$, $\mathbf{M}_{i} \sim P(\mathbf{M}), i=1,\ldots,N_f$. The output of the AE for any $\mathbf{M}_{i} \in D_m$ is
\begin{equation}
    \widehat{\mathbf{M}_i} = f_{t}(\mathbf{M}_i,{{{\boldsymbol \theta}_{t}}}).
\end{equation}
Since the AE is trained for task $T_{t}$ and as $\mathbf{M}_{i} \in D_m$ is sampled from another distribution, $\widehat{\mathbf{M}_i}$ will be a meaningless reconstruction of $\mathbf{M}_i$. Alternatively, it can be said that the activations obtained on passing $\mathbf{M}_i \sim P(\mathbf{M})$ through the trained $f_{t}(\cdot,{{{\boldsymbol \theta}_{t}}})$ does not align with $T_{t}$. This can be observed from the MAE between $\mathbf{M}_i$ and $\widehat{\mathbf{M}_i}$, and also from the t-SNE representations of the bottleneck layer (latent space) for several samples drawn from $P(\mathbf{M})$ (Figure \ref{tsne_main}: Iteration $r=0,1$). 
% The reconstruction MAE between successive iterations drop, and the clusters formed by $D_m \sim D_{T}$ with increasing iterations.
% Fig \ref{flash_withMAE} (second column) and 
% It is important to note that as ${{{\boldsymbol \theta}_{t-1}}}$ inherently captures $P(\mathbf{x}), \forall \mathbf{x} \in D_{T_{t-1}}$, $\widehat{\mathbf{M}}'$ is expected to be drawn towards $P(\mathbf{x})$, through ${{{\boldsymbol \theta}_{t-1}}}$.

\begin{figure}[htp]
	\centering
	\includegraphics[width=0.4\textwidth]{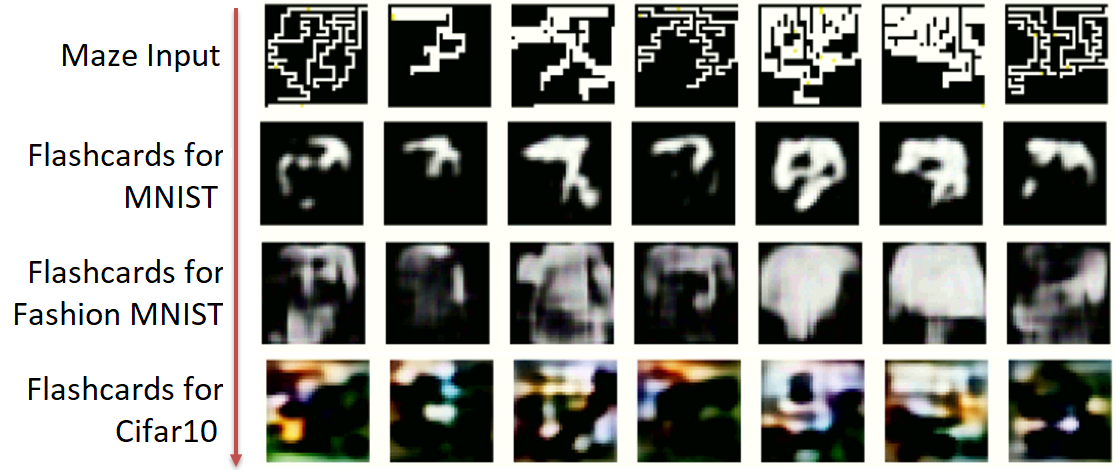}
	\caption{Transformation from maze patterns into Flashcards for different tasks: MNIST, Fashion MNIST and Cifar10.}
	\label{flashcards_change_across_tasks}
	\vspace{-0.3cm}
\end{figure}

\begin{figure*}[!t]
	\centering
	\includegraphics[width=1.0\textwidth]{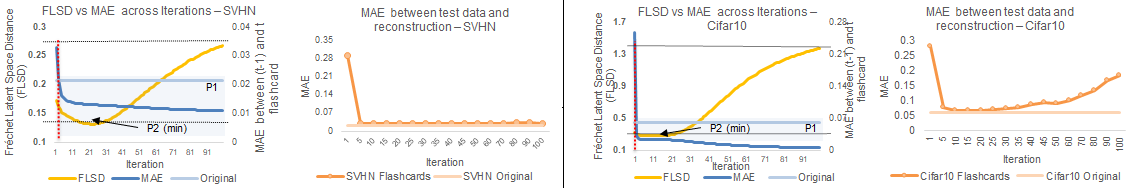}
	\caption{Analysis on $r$ for flashcards construction, demonstrated using SVHN and Cifar10. Odd columns portray metrics corresponding to change on passing multiple random maze inputs in iterative fashion through an AE. Orange curve depicts the trend of Fr\'echet latent space distance (FLSD) between intermediate input at (t) iteration and original data samples, dark blue curve shows constant decrease in reconstruction error between (t-1) and (t), across different iterations for each dataset. Light blue line serves as reference MAE on original data. Property P1 is satisfied when the blue curve goes below $\epsilon_1$ (marked by dotted red line) and P2 is satisfied at the first minima of FLSD, which occurs around $10 \leq r \leq 20$, irrespective of dataset. Even columns compare AE error (MAE) between test data and its reconstruction, when trained on flashcards constructed from different iterations. This again confirms the acceptable range of $r$ and that flashcards construction is {\em not critically sensitive} to the choice of $r$ within the range.}

\label{flsd_mae}
\vspace{-0.25cm}
\end{figure*}

Let $\mathbf{F}_{ri}$ be the output after a series of $r$ \emph{recursive iterations/passes} of $\mathbf{M}_i$ through the trained AE ($f_{t}(\cdot, {\boldsymbol \theta}_{t})$), i.e., 
\begin{equation}
   \mathbf{F}_{ri} =  f_{t}^{r}(f_{t}^{r-1}(\cdots f_{t}^{1}(\mathbf{M}_i,{{{\boldsymbol \theta}_{t}}})\cdots)). \label{eq_recur}
\end{equation}
Such recursive passing (output $\rightarrow$ input $\rightarrow$ output) gradually attunes the input to  ${\boldsymbol \theta}_{t}$. This can be observed in Figure \ref{tsne_main} where the MAE reduces over recursive iterations, and the latent space representations of samples drawn from $P(\mathbf{M})$ increasingly overlap with those from $P(\mathbf{X})$. Let %As the recursive iterations (output $\rightarrow$ input $\rightarrow$ output) progresses, the random images gets accustomed to $f_{t}(\cdot)$ and hence their reconstruction error over successive iterations reduces. In other words,
\begin{equation}
  \gamma_1 \leq | \mathbf{F}_{ri} -\mathbf{F}_{(r-1)i} | \leq \gamma_2,~\forall  \mathbf{M}_i \in D_{m} \sim P(\mathbf{M}), \label{flash_recon}
\end{equation}
where $\gamma_1$, and $\gamma_2$ are the lower and upper bounds of reconstruction error for any $\mathbf{M}_i \in D_{m}$. The aim now is to have a collection set $F_t =\{\mathbf{F}_{ri}: i=1,\ldots,N_f\}$ for task $T_t$ such that the following properties are satisfied:
\begin{itemize}
\item P1: $\gamma_2 < \epsilon_1$ 
% and $\varepsilon_2 < \epsilon_1$ 
for $\forall \mathbf{M}_i \in D_m$, %(both bounds of reconstruction error is less than the lower bound of AE) 
\item P2: The latent space distribution of $F_t \sim D_{t}$, i.e. $~D_m \rightarrow D_t$. 
\end{itemize}
Such a set of constructions ($F_t$) obtained through recursive iterations of random inputs (drawn from $P(\mathbf{M})$) are defined as {\em flashcards}. 
% In this regard, we introduce 
% These {\em flashcards} have the ability to capture knowledge representations from ${{{\boldsymbol \theta}_t}}$ through presenting random patterns recursively to the trained AE.
Such constructed flashcards are expected to capture the knowledge from the trained network and serve as a potential alternative to $D_{t}$. Figure \ref{flashcards_change_across_tasks} illustrates a few flashcards constructed with same set of random images, from AEs trained with different datasets. Although these flashcards are intended to capture the knowledge representations in ${{{\boldsymbol \theta}_{t}}}$ as a function of $\mathbf{M}_i \in D_{m}$, it can be observed from Figure \ref{flashcards_change_across_tasks} that they do not bear direct shape similarities with their respective datasets. In fact, during the recursive process, the raw input image patterns are modified/transformed to textures that are suited to the trained AE model. This is in-line with \citep{TEXTURE_ICLR_2019} which shows that neural networks learn predominantly texture than shape by default. Now, since an AE is trained to maximize $\log P(\mathbf{X}| {{{\boldsymbol \theta}_{t}}})$, the following fact holds:
% Therefore, due to the training characteristic, flashcards effectively capture the texture, but not the shape as effectively. 
%Empirical evidence shows that $r\geq 5$ achieves the desired properties (more details in Section 4).
% However, whether a network specifically learns the shape is dependent on the specific loss function used. In our work, we do not employ special methods , rather use the general loss as generally used. Since the network does not inherently learn the shape, we strengthen the initial input using certain patterns % However, these recursively constructed flashcards are fine-tuned to align with network parameters trained with $D_{T}$. 

\begin{fac}
Let $D_{A}$ and $D_{B}$ be two i.i.d. datasets whose elements are drawn from $P(\mathbf{X})$. If there exists a trained AE $f_{A}(\cdot, {\boldsymbol \theta}_{A})$, to reconstruct $D_{A}$ with error $[\beta_1, \beta_2]$, $\forall \mathbf{X} \in D_{A}$, then the AE $f_{A}(\cdot, {\boldsymbol \theta}_{A})$ can reconstruct $D_{B}$ with the same error range. 
\label{fact1}
% \vspace{-0.1cm}
\end{fac}
The converse is not true because any image dataset $D_{B}$ with images that have few pixels with non-zero values (other pixels as zeros), can still yield smaller reconstruction error and not necessary to be drawn from $P(\mathbf{X})$ (satisfying P1). Hence, both P1 and P2 for $F_t$ need to be satisfied. As P1 and P2 are dependent on the initial input distribution $P(\mathbf{M})$, it is important to have a suitable $P(\mathbf{M})$ to get $F_t$ ({$P(\mathbf{M}) = P(\mathbf{X})$ is one trivial and uninteresting solution}). As finding  a perfect $P(\mathbf{M})$ is an ideal research problem by itself, in this work, we use random maze pattern images as a potential candidate $D_m$ to get $F_t$. The above leads to the following hypothesis:
\begin{hyp}
\label{hypothesis1}
Since the AE trained on $D_{t}$ also reconstructs elements from $F_t$ (Property P1) and has matching latent space distribution (Property P2), flashcards ($F_t$) can be used as a set of pseudo-samples for $D_{t}$ to learn $f_{t}(\cdot, {{{\boldsymbol \theta}_{t}}})$, when trained from scratch using only $F_t$. 
\end{hyp}
% \vspace{-0.1cm}

\subsection{Flashcard Construction}
% \vspace{-0.1cm}
\begin{algorithm}
\small
\caption{Flashcard construction}
\label{algo: flashcard_creation}
\begin{algorithmic}[1]
% \REQUIRE Input image dimensions: $m, n, c$ :  width, height and number of channels
% \REQUIRE $f_{t-1}(\cdot)$: AE model for {$t-1$^{th}} task $T_{t-1}$
\REQUIRE $f_{t}(\cdot, {\boldsymbol \theta}_{t})$: AE model for task $T_{t}$
\REQUIRE $N_f$: Number of flashcards 
\REQUIRE $r$: Number of recursive iterations through $f_{t}(\cdot, {\boldsymbol \theta}_{t})$at
% \REQUIRE $f_{t}(\cdot, {\boldsymbol \theta}_{t})$: AE model for task $T_{t}$, $N_f$: Number of flashcards, $r$: Number of recursive iterations through $f_{t}(\cdot, {\boldsymbol \theta}_{t})$ 

\STATE Let $D_{m} \leftarrow \left\{\mathbf{M}_{1},\ldots\mathbf{M}_{N_f}\right\}$, where $\mathbf{M}_{i} \sim P(\mathbf{M}), \forall i$, $P(\mathbf{M})$ corresponds to the distribution of initial input patterns
% \STATE maze\_patterns $\leftarrow$ Obtain $n_f$ initial maze patterns 
\STATE  $F_{t} \leftarrow \{\}$, where $F_{t}$ denotes the set of flashcards
\FOR{$i \in \left\{1, \ldots, N_f\right\}$}
    % \STATE Pass the random\_images list to $f_{\theta}$ (AE model) and predict reconstructions
   \STATE $\mathbf{F}_{ri} =  f_{t}^{r}(f_{t}^{r-1}(\cdots f_{t}^{1}(\mathbf{M}_i,{{{\boldsymbol \theta}_{t}}})\cdots))$
   \STATE $F_{t} = F_{t} \cup \{\mathbf{F}_{ri}\}$ 
    %   \STATE Pass $iter$ times: $f_{t-1}({\rm random\_images}[k], {{{\boldsymbol \theta}_{t-1}}})$ 
    %   \STATE Overwrite the reconstructions as the random\_images
    % \ENDFOR
\ENDFOR 
% \RETURN random\_images (Flashcards)
\RETURN $F_{t}$
\end{algorithmic}
\vspace{-0.1cm}
\end{algorithm}

% \subsection{Hyperparameters}
% For flashcards construction, the raw input and number of iterations are required. 

Algorithm \ref{algo: flashcard_creation} provides the steps to construct flashcards using a trained AE model. For flashcards construction, the following hyperparameters are useful - (i) number of recursive iterations ($r$), (ii) number of flashcards to construct ($N_f$), (iii) choice of initial input distribution ($P(\mathbf{M})$). Figure \ref{flsd_mae} portrays empirical analysis of selecting the optimal iteration, considering complex (with background) datasets such as SVHN and Cifar10 dataset. Figure \ref{flsd_mae} (columns 1,3) shows Frechet Latent Space Distance (FLSD is inspired from FID \citep{heusel2017gans} and calculated as the Frechet distance between latent space activations of flashcards versus original samples (smaller the better), from the eyes of the trained AE) (Orange) and MAE (Blue), as a function of $r$. Initially, increase in $r$ leads to decrease in FLSD, up to certain iterations, this indicates that flashcards distribution is getting closer to learned distribution. As $r$ increases further, though MAE keeps going down, FLSD increases, because propagation of reconstruction error in AE causes drift in the features. Based on the empirical analysis $10\leq r\leq 20$ is a sensible choice that worked well across all of our experiments.
% optimal for flashcard generation, irrespective of the datasets. 
Also, reconstruction error (Figure \ref{flsd_mae}, columns 2,4) is observed to be low in this iteration range. It further indicates that the flashcards are not critically sensitive to $r$. Hence, in all the experiments we set $r$ to be 10.

\begin{table}
\small
\centering
{ 
\begin{tabular}{ p{0.7cm} p{0.9cm} p{0.9cm} p{1.2cm} p{1.3cm} p{1cm}} \hline %|c|c|c|c|
{\textbf{Dataset}} & \textbf{Original MAE} & \textbf{Flash-cards MAE} & \textbf{Alpha on Original} & \textbf{Alpha on Flashcards} & \textbf{Alpha on Untrained}\\\hline
{MNIST} &  $0.0184$ & $0.0491$ & $1.9185$ & $2.0531$ & 1.4343 \\ 
{Fashion} &  $0.0259$ & $0.0440$ & $2.0044$ & $2.1158$ & 1.4343\\ 
{Cifar10} &  $0.0564$ & $0.0686$ & $2.0266$ & $2.1531$ & 1.4343 \\ \hline
\end{tabular}
\caption{Reconstructions from two separate AE, first trained on original data, and second trained on flashcards.The metrics indicate that flashcards are sufficient alternatives towards learning the particular data. Weighted Alpha \citep{martin2020heavy} is based  on  HT-SR  Theory. Alpha values are closer for the trained networks which indicates similarity between the two network weights. Reported values are obtained by taking mean over 5 runs.}
\label{table:scratch-original}
}
\end{table}
% \vspace{-0.05cm}

% \begin{figure}[h]
% 	\centering
% 	\includegraphics[width=0.49\textwidth]{figure/flashcard-alt-org-recon.png}
% 	\caption{Reconstructions from two separate networks, first trained on original data, and second trained on flashcards. The reconstructions for both the networks are visually similar.}
% 	\label{fig:scratch-original}
% \end{figure}

\begin{table}
\small
\centering
{ 
\begin{tabular}{ c p{1cm} p{1.2cm} p{0.95cm} p{0.95cm}} \hline %|c|c|c|c|
% \multirow{2}{*}{\textbf{Arch. Type}} & \multirow{2}{*}{ Params} & \multirow{2}{*}{\textbf{Original}} & \multirow{2}{*}{ \begin{tabular}{@{}c@{}} \textbf{AE1} \\ \textbf{} \end{tabular}} & \multirow{2}{*}{ \begin{tabular}{@{}c@{}} \textbf{Flashcards} \\ \textbf{from AE2} \end{tabular}} \\ {} & {} & {} & {} & {} \\\hline
{\textbf{Train Arch. Type}} & {\textbf{Params}} & {\textbf{Test on Original}} & {\textbf{Test on AE1}} & {\textbf{Test on AE2}} \\\hline
{AE1 - Variant I} &  $94,243$ & $0.0640$ & $0.0725$ & $-$\\ 
{AE2 - Smaller arch.} &  $24,083$& $0.0787$ & $0.0939$ & $0.0963$\\ 
{AE2 - Larger arch.} &  $372,803$ & $0.0512$ & $0.0681$ & $0.0570$\\\hline
{AE1 - Variant II} &  $298,947$ & $0.0437$ & $0.0599$ & $-$\\ 
{AE2 - Smaller arch.} &  $24,083$& $0.0358$ & $0.0385$ & $0.0389$\\ 
{AE2 - Larger arch.} &  $372,803$ & $0.0512$ & $0.0654$ & $0.0570$\\\hline
\end{tabular}
\caption{Building autoencoder AE2 using flashcards obtained from AE1 (two variants) trained on Cifar10 dataset. Columns AE1 and AE2 refer to test MAE on Cifar10 when flashcards obtained from AE1 and AE2 are used to train for reconstruction, respectively. Better performance on AE2 Smaller using flashcards from AE1 shows transferability using existing flashcards.
% We observe that the AE2 Smaller network for case 2 is able to achieve lower MAE than that of originally trained AE1.
}
\label{table:scratch-architecture}
}
% \vspace{-0.3cm}
\end{table}

% For simpler datasets, a well trained AE reconstruction error on original data is less, implying that irreducible error (noise) is considerably less. Repeated flashcard passes through such AE do not propagate noise to a great extent, and do not cause deterioration in reconstructions across iterations. For complex datasets like Cifar10, irreducible error after training adds noise with each iteration, and causes the captured patterns to get deteriorated / smoothed out. Due to this effect, magnitude of FLSD starts increasing with iterations (P1 is still satisfied but P2 not anymore). From the figure, FLSD between 10-100 iterations (y-axis left) is ~0.03 for Fashion MNIST. Test error at bottom is also relatively flat after few iterations. For SVHN, initially it appears that FLSD trend has increased, but magnitude is comparatively less ~0.10, and MAE between flashcard iterations (y-axis right) has flattened at ~0.01, thereby no change to test error at bottom. For Cifar10, due to deterioration across iterations, FLSD is consistently increasing, and test error is also increasing at bottom. Repeated passing beyond this point did indeed reduce the error, but it 
% deteriorated the features captured. 

%(Visualization seen in Figure \ref{tsne_main}) 

In general, as more flashcards 
% and coreset exemplars 
are introduced, performance gets better. In our experiments, number of flashcards constructed equals the same no. of samples in original data.\footnote{Improvement with increase in flashcards in Supplementary} Note from \citep{ostapenko2019learning}, that memory requirement is calculated based on network parameters and is independent of generated samples.

% Figure \ref{flash_card_samples_increase} shows improvement in performance as more flashcards and coreset exemplars are introduced (MNIST and Cifar10 example). Generally, for scratch setting, number of flashcards constructed equals the same number of samples in original data. For CL setting, 10\% (5000) flashcards constructed before training a new task (irrespective of number of previous tasks) gave fair results. Increasing this number helps to remember previous tasks, similar to episodic memory, but it also increases the regularization effect. Note from \citep{ostapenko2019learning}, the memory requirement is calculated based on network parameters and is independent of generated samples.

As mentioned above, the choice of $P(\mathbf{M})$ is the maze-like random image patterns that resembles edges and shapes. 
% Finding the optimal $P(\mathbf{M})$ is a research question in itself. 
Gaussian random noise (pixel-wise \& geometric patterns) were also considered and it was observed that pixel-wise noise does not help in capturing representations, and
% remembering previous tasks, 
maze patterns outperformed geometric patterns.
% were efficient only to a smaller extent.

\subsection{Flashcards for Capturing Representations}
\label{flashcards-capture-representation}
We first verify Hypothesis \ref{hypothesis1} by demonstrating the efficacy of flashcards to capture representations on single-task scenario (Algorithm in Supplementary). %train network parameter for single task cases (scratch case)
% We later extend the property of flashcards to CL scenario in Section \ref{perf}.
% Algorithm for single-task scenario is present in Suppl. 

\noindent \textbf{Alternative to Original Dataset:}
Consider three different datasets: MNIST, Fashion MNIST, and Cifar10. Let 3 autoencoders AE1, AE2, AE3 (all with same architecture) be trained for each of these 3 datasets separately using Eqn. \eqref{AE} as loss function. 
% Few samples of original and reconstructed images for each of these datasets, along with the corresponding MAEs are shown in Figure \ref{flashcards_change_across_tasks}. 
Flashcards ($F_t$) constructed from each of these AEs, for the same set of random input maze pattern images ($D_m$) 
(few samples shown in  Figure \ref{flashcards_change_across_tasks}) 
are used respectively to train new AE models from scratch (with the same architecture), namely AE-flash1, AE-flash2, and AE-flash3. Results in Table \ref{table:scratch-original} show  reconstructions are very close to AE trained on original. These results confirm the hypothesis that flashcards indeed capture network parameters as a function of $D_m$, therefore be used as pseudo-samples / training data (as network parameters involved in flashcard construction are initially learned by training with the original dataset $D_{t}$).

\noindent \textbf{Knowledge Distillation to smaller/larger architecture:} Consider training a network AE1 with a certain data $D_{T_1}$. Once trained, $D_{T_1}$ may not be available because of confidentiality/privacy, storage requirements, etc. In regular KD, this creates a bottleneck because original data is still required to be shown to the new network. If in the future, there is a newer and better architecture AE2, our method enables migration by training on flashcards constructed from AE1. We show this is possible by training two network variants - one smaller and one larger modified architectures than original
% Though the gain (or loss) in performance is not so high, 
% results summarized in 
(Table \ref{table:scratch-architecture}).

\subsection{Flashcards for Replay in CL}
% \vspace{-0.4cm}
\label{flashcards-replay-in-cl}
% \vspace{-0.4cm}
% \begin{algorithm}
% \caption{Flashcard for Replay in CL}
% \label{algo: flashcard_CL_case}
% \begin{algorithmic}
% \REQUIRE $f_{t}(\cdot, {\boldsymbol \theta}_{t})$: Autoencoder model for task $T_{t}$
% \REQUIRE $N_f$: Number of flashcards
% \REQUIRE $r$: Number of iterations through $f_{t}(\cdot, {\boldsymbol \theta}_{t})$ %over the autoencoder model
% \REQUIRE $D_{t+1}$: Data for task $T_{t+1}$
% \STATE $F_{t} \leftarrow$ \textbf{Algorithm \ref{algo: flashcard_creation}} ($f_{t}(\cdot, {\boldsymbol \theta}_{t}), N_f, r$), where $F_{t}$ is the set of flashcards for task $T_t$
% \STATE Train a new autoencoder model $f_{t+1}(., {\boldsymbol \theta}_{t+1})$ on flashcards $F_t$ and data $D_{t+1}$
% \STATE Initialize ${\boldsymbol \theta}_{t+1} \leftarrow {\boldsymbol \theta}_{t}$
% \begin{eqnarray}
%     L_{task}({\boldsymbol \theta}_{t+1}) = \frac{1}{|D_{t+1}|} \sum_{n=1}^{|D_{t+1}|} |D_{t+1}^{(n)} - f_{t+1}(D_{t+1}^{(n)}, {\boldsymbol \theta}_{t+1})|
%     \nonumber
% \end{eqnarray}
% \begin{eqnarray}
%     L_{flashcards}({\boldsymbol \theta}_{t+1}) = \frac{1}{N_f} \sum_{n=1}^{N_f} |F_{t}^{(n)} - f_{t+1}(F_{t}^{(n)}, {\boldsymbol \theta}_{t+1})|
%     \nonumber
% \end{eqnarray}
% \begin{eqnarray}
%     {\boldsymbol \theta}_{t+1}^{*} \leftarrow \argmin_{{\boldsymbol \theta}_{t+1}} {\big (} L_{task}({\boldsymbol \theta}_{t+1}) + L_{flashcards}({\boldsymbol \theta}_{t+1}) {\big )}
%     \nonumber
% \end{eqnarray}
% \RETURN $f_{t+1}(., {\boldsymbol \theta}_{t+1}^{*})$
% \end{algorithmic}
% \end{algorithm}
Consider a sequence of $T$ tasks \{$T_1,\ldots,T_t,\ldots,T_T$\}.  In the CL scenario, the AE for task $T_{t+1}$ is required to be trained on top of previous learned tasks $T_1,\ldots,T_{t}$. In other words, ${{{\boldsymbol \theta}_{t+1}}}$ is adapted from the previously trained network parameters ${{{\boldsymbol \theta}_{t}}}$. Such AE training for task $T_{t+1}$ may result in AE forgetting the representations learned till the previous task $T_{t}$. Unlike other CL based approaches which aims to preserve ${{{\boldsymbol \theta}_{t}}}$ through regularization, memory replay (either episodic or using external generative networks), architectural strategies, or their combinations, we use the flashcards constructed on ${{{\boldsymbol \theta}_{t}}}$ along with data for task $T_{t+1}$, while training for task $T_{t}$.
% It is worth to note that \citep{sodhani2020toward} lists four desirable (not mandatory) properties for a CL model.
% Flashcards satisfy properties desirable for CL - (i) Knowledge Retention, (ii) Knowledge Transfer, and, (iii) Parameter Efficiency.
Flashcards are required to be constructed only at the end of task $T_{t}$ {\em irrespective} of the number of preceding tasks, so that knowledge representations for tasks $\{1, \ldots, t\}$ can be captured and trained with the next task $T_{t+1}$. Thus, the proposed method avoids storing of flashcards for each successive task, thereby significantly reducing the memory overhead, while ensuring robust performance (Section \ref{perf}. Algorithm \ref{algo: flashcard_CL_case} provides the steps for using flashcards as a replay strategy for CL. In our CL experiments, flashcards 10\% (of next task samples) were constructed on-the-fly before training a new task (irrespective of number of previous tasks). Increasing this number helps to remember previous tasks better.

\begin{algorithm}
\small
\caption{Flashcard for Replay in CL}
\label{algo: flashcard_CL_case}
\begin{algorithmic}[1]
\REQUIRE $f_{t}(\cdot, {\boldsymbol \theta}_{t})$: AE model for task $T_{t}$; $N_f$: Number of flashcards; $r$: Number of iterations through $f_{t}(\cdot, {\boldsymbol \theta}_{t})$ 
\REQUIRE $D_{t+1}$: Data for task $T_{t+1}$
\STATE $F_{t} \leftarrow$ \textbf{Algorithm \ref{algo: flashcard_creation}} ($f_{t}(\cdot, {\boldsymbol \theta}_{t}), N_f, r$), where $F_{t}$ is the set of flashcards for task $T_t$
\STATE Train the autoencoder model $f_{t+1}(., {\boldsymbol \theta}_{t+1})$ on flashcards $F_t$ and data $D_{t+1}$
\STATE Initialize ${\boldsymbol \theta}_{t+1} \leftarrow {\boldsymbol \theta}_{t}$
% \begin{eqnarray}
%     L_{task}({\boldsymbol \theta}_{t+1}) = \frac{1}{|D_{t+1}|} \sum_{n=1}^{|D_{t+1}|} |D_{t+1}^{(n)} - f_{t+1}(D_{t+1}^{(n)}, {\boldsymbol \theta}_{t+1})|
%     \nonumber
% \end{eqnarray}
% \begin{eqnarray}
%     L_{flashcards}({\boldsymbol \theta}_{t+1}) = \frac{1}{N_f} \sum_{n=1}^{N_f} |F_{t}^{(n)} - f_{t+1}(F_{t}^{(n)}, {\boldsymbol \theta}_{t+1})|
%     \nonumber
%  \vspace{-0.2cm}
 \STATE Optimize ${\boldsymbol \theta}_{t+1}$ using given loss  
\begin{eqnarray}
    {\boldsymbol \theta}_{t+1}^{*} \leftarrow  \argmin_{{\boldsymbol \theta}_{t+1}} {\big (} \frac{1}{|D_{t+1}|} \sum_{n=1}^{|D_{t+1}|} |D_{t+1}^{(n)} - f_{t+1}(D_{t+1}^{(n)}, {\boldsymbol \theta}_{t+1})| \nonumber \\ 
   + \frac{\lambda}{N_f} \sum_{n=1}^{N_f} |F_{t}^{(n)} - f_{t+1}(F_{t}^{(n)}, {\boldsymbol \theta}_{t+1})| {\big )}
    \nonumber
\end{eqnarray}
\vspace{-0.35cm}
\RETURN $f_{t+1}(., {\boldsymbol \theta}_{t+1}^{*})$
\end{algorithmic}
\end{algorithm}

\section{Related Works} \label{relworks}

In general, CL algorithms are based on  architectural strategies, regularization, memory replay, and their combinations \citep{clreview}. By the intrinsic nature of flashcards (as discussed in Sections \ref{flashcards-capture-representation} and \ref{flashcards-replay-in-cl}), they implicitly exhibit the characteristics of both regularization (refer Algorithm \ref{algo: flashcard_CL_case} point 4) and replay, and will therefore be compared with the respective CL strategies. 

Regularization strategies such as \citep{EWC2017}, \citep{si} learn new tasks while imposing constraints on the network parameters to avoid straying too much from those learned from the previous tasks. \citep{lwf} uses current task samples to regularize the soft labels of past tasks, thus avoiding explicit data storage. However, it is important to note that these regularization methods perform well in homogeneous task environments, where it is possible to find mutually sub-optimal points in the solution space. However, such methods suffer when subsequent tasks are from different domains / heterogeneous datasets (also reported in \citep{lwf}).

% In the CL setting, flashcard serves as both regularization and replay to   
% \citep{DEN2018}\citep{PNN} \citep{NIPS2019_9518}, regularization \citep{GEM}, \citep{chaudhry2018efficient}, \citep{resiaaai2020} \citep{Ebrahimi2020Uncertainty-guided} are prevalent, algorithms based on rehearsal or replay strategies are recently gaining momentum. 

In a rehearsal mechanism,
% the neural network is "reminded" of the previous tasks, with their knowledge representatives, at the end of every task training. For example, 
subset of samples from previous tasks, referred to as coreset samples / exemplars, serve as memory replay \citep{GEM}, \citep{chaudhry2018efficient}, \citep{castro2018end}.
% \citep{vcl}.
However, they suffer from excessive storage requirements that grows with the number of tasks. On the other hand, generative replay approaches \citep{braininspired2020}, \citep{GCCLreplay_AAAI2020},
\citep{onlinecont_Maxreplay_Neurips2019}, \citep{ICLOGMReplay_TNNLS2020} face challenges in scalability to complicated problems with many tasks, since they
% For instance, GAN based replay methods 
involve heavy training and computation associated with an auxiliary network to generate visually meaningful images for each task. 
% (plus extra effort to mitigate issues such as mode collapse and training instability).

Approaches discussed above are either suitable only for homogeneous data, or are memory intensive as they involve {\em preserving} samples, or computationally intensive in "generating" samples, or requiring task identifiers. The proposed flashcards based capturing of learned representations are constructed on the fly, requires only simple autoencoder, independent of the no. of tasks, and helpful {\em both} as regularization and replay mechanism. Flashcards can be used as {\em pseudo samples} constructed with low computational and memory expense. Furthermore, flashcards can scale across tasks without considerable drop in performance. Some main characteristics of flashcards against other replay based CL approaches are summarized in Table 1 of Supplementary. 

\vspace{-0.15cm}
\section{Experiments in Continual Learning}% 
\label{perf}
\vspace{-0.10cm}

We provide experimental validation of using flashcards in CL scenario as replay for variety of different applications such as: (i) continual reconstruction (Section 4.1), (ii) continual denoising (Section 4.2), and continual classification (Section 4.3), split into Task Incremental Learning (Task-IL) and New Instance Learning (NIL). For the first two experiments, we use 5 heterogeneous public datasets (referred as Sequence5), namely MNIST, Fashion MNIST, Cifar10, SVHN, and Omniglot.
% , and evaluate the ability of flashcards to represent the datasets when trained sequentially without forgetting the earlier tasks.
For Task-IL classification, we use Sequence3 comprising of Cifar10, MNIST and Fashion MNIST. For NIL classification, we use Cifar10.
% , and introduce 5 sessions with new instances to learn per session. 

For reconstruction, denoising and NIL classification, flashcards are constructed using an AE architecture with 4 layers of down/upsampling and 64 conv filters per layer. The bottleneck dim 256 is 12x reduction from image space, and serves to demonstrate effectively the forgetting in CL. The architecture is a simplified variant of the VGG style
{\footnote {More details on arch. and hyperparameters in Supplementary.}}
, adapted with lesser layers and filters to enable proper reconstruction for the baseline. All hidden layers employ tanh activation. The network is trained for 100 epochs per task, and optimized using Adam with a fixed learning rate of 1e-3. Each minibatch update is based on equal number of flashcards and the current task samples. 10\% of the training data is allocated for validation. Early stopping is employed if there is no improvement for over 20 epochs. 5K flashcards with $\lambda=1$ was used as replay for reconstruction and denoising. The workstation used for training has 64GB RAM and NVIDIA RTX Titan GPU. For all experiments, we compare with upperbound (Joint Training), lowerbound (Sequential Fine-tuning, SFT), and examples of methods using regularization, generative and episodic replay as baseline.

\begin{table}
% 	\begin{minipage}{0.5\linewidth}
		\centering
		\small
		\begin{tabular}{p{1.7cm} p{0.4cm} p{0.4cm} p{0.4cm} p{0.52cm} p{0.55cm} p{0.67cm} p{0.67cm}} \hline %|c|c|c|c|
        % |p{1.5cm}|p{0.5cm}|p{1cm}|p{1cm}|p{1cm}|p{1.15cm}| } \hline %|c|c|c|c|

        \textbf{Method} &\textbf{Type} & \textbf{Nw. (MB)} & \textbf{Mem. (MB)} &  \textbf{Recon MAE} & \textbf{Recon BWT} & \textbf{Denoise MAE} & \textbf{Denoise BWT} \\ \hline
        {JT} & - & 1.5 & 798.7 & $0.035$ & {-} & $0.055$ & {-} \\ 
        {SFT} & - & 1.5 & {-} & $0.552$ & $-0.64$ & $0.547$ & $-0.63$\\ 
        {Coreset 500} & ER & 1.5 & 1.5 & $0.058$ & $-0.02$ & $0.075$ & $-0.02$ \\ 
        {Coreset 5K} & ER & 1.5 & 15.4 & $0.049$ & $-0.02$ & $0.059$ & $-0.02$ \\ \hline
        {LwF} & Reg & 1.5 & {-} & $0.549$ & $-0.52$ & $0.437$ & $-0.33$ \\ 
        {VAE-5K+AE} & GR & 2.9 & {-} & $0.075$ & $-0.06$ & $0.071$ & $-0.04$ \\ 
        {CL-VAE 5K} & GR & 1.4 & {-} & $0.352$ & $-0.16$ & $0.546$ & $-0.28$ \\ 
        \textbf{Flashcards5K} & FR &  $\mathbf{1.5}$ & {-} & {$\mathbf{0.053}$} & $\mathbf{-0.03}$ & $\mathbf{0.058}$ & $\mathbf{-0.03}$\\ \hline
        \end{tabular}
        \vspace{-0.1cm}
        \caption{Continual learning for reconstruction and denoising (added standard noise of 0.1), values averaged over 5 runs (stdev in Supplementary). Comparison in terms of average error (MAE), forgetting (BWT), Network Capacity (Nw. (MB)) and External Memory (Mem. (MB)) is presented. Flashcards performs better than generative replay, and is on-par with episodic memory (coreset), without need for external memory. Reg=Regularization, ER=Episodic Replay, GR=Generative Replay, FR=Flashcard Replay.}
		\label{table::continual_recon-short}
		\end{table}
% 	\end{minipage}
% 	\begin{minipage}{0.475\linewidth}
\begin{figure}
		\centering
		\vspace{-0.2cm}
		\includegraphics[width=0.48\textwidth]{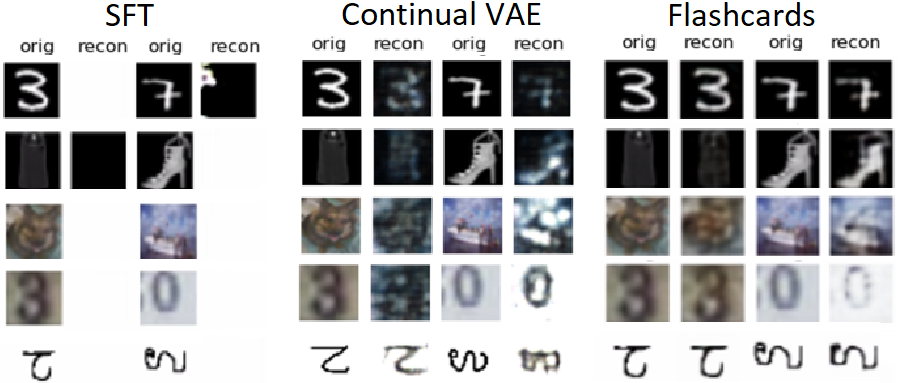}
		\vspace{-0.5cm}
		\captionof{figure}{Comparison between different methods for continual reconstruction using Sequence5. The figure shows the impact of forgetting after the end of the {\em last} task, i.e. Omniglot (all intermediate results are in Supplementary). Each row presents a dataset/task. Sequential Fine Tuning (SFT, left), is not able to reconstruct, even when presented with Omniglot, because (i) catastrophic forgetting and (ii) unsuitable transfer learned weight initialization. Continual VAE (center) fares better but has artifacts and color loss in the reconstructions. Flashcards (right) remembers all previous tasks.}
		\label{seqof5_last_task}
% 	\end{minipage}
\vspace{-0.25cm}
\end{figure}

\begin{figure}[!htpb]
\centering
\includegraphics[width=0.44\textwidth]{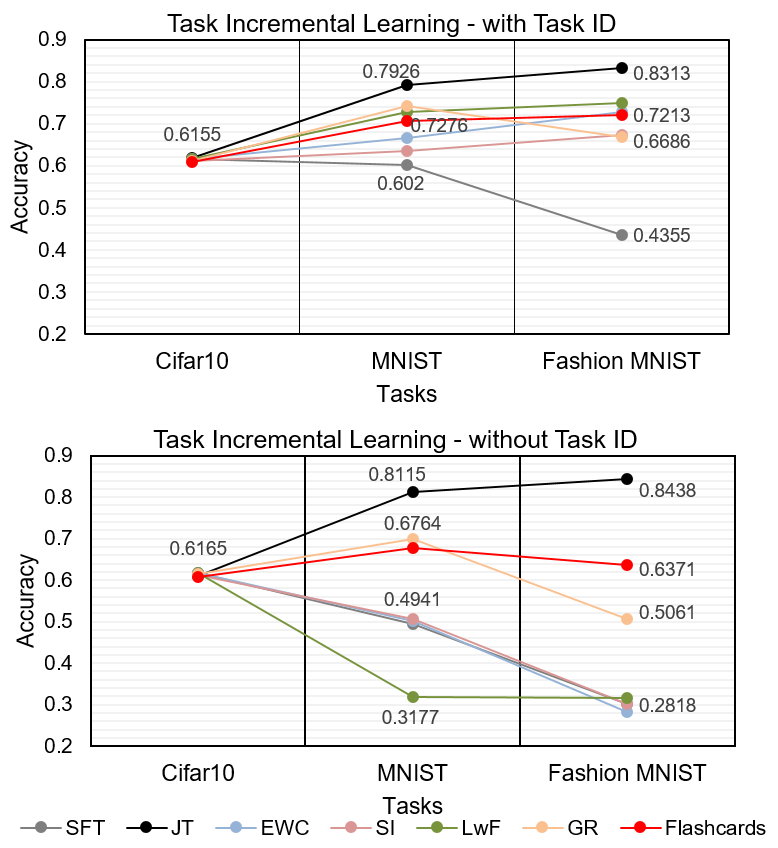}
\vspace{-0.35cm}
\caption{Comparison of different methods for task incremental learning using Sequence3 tasks (Cifar10-MNIST-FashionMNIST). When task id is provided, flashcards performs similar to baseline methods. However, baselines fail in absence of identifier. Flashcards is robust to degradation across different domain tasks and outperforms other methods. Comparison: upperbound (JT), lowerbound (SFT), EWC, SI, LwF and Generative Replay with VAE (GR) accuracy. Results are averaged over 5 runs.}
\label{taskIL-with-without-ids}
\vspace{-0.3cm}
\end{figure}
\subsection{Continual Reconstruction}

% \begin{figure*}
% 	\centering
% 	\includegraphics[width=0.68\textwidth]{LaTeX/figure/results_recon.png}
% 	\caption{Results (need to add appropriate caption).}
% 	\label{results_recon}
% \end{figure*}

Sequence5 is used to demonstrate reconstruction in CL. We measure individual task MAE and 
% before and after observing the data for the given task as well as the ability of transferring knowledge from one task to another, by measuring 
BWT \citep{GEM}, customized for reconstruction. Results are listed in Table \ref{table::continual_recon-short} and visualized in Figure \ref{seqof5_last_task}.
% and Figure \ref{results_recon_graph}. 
Comparison with baseline is summarized in the captions for ease of convenience. To the best of our knowledge, we are the first to report continual reconstruction. We compare flashcards against different replay and regularization techniques and observe it outperforms most baselines and on-par with episodic replay. Results for permutations of Seq3 (MNIST-Fashion-Cifar10) in Suppl.

\subsection{Continual Denoising}

A more challenging extension to traditional AE reconstruction is denoising. We impose noise sampled from a standard normal distribution factored by 0.1 to Sequence5, and expect the network to denoise it. To the best of our knowledge, we are the first to report Continual Denoising. 
% We omit VAE based replay since flashcards outperform them in reconstruction itself.
Results reported for Denoise MAE and BWT in Table  \ref{table::continual_recon-short} are obtained by adding noise factor of 0.1 to original images. Flashcards perform better compared to baseline approaches. More visual results for different noise levels are provided in Supplementary.

% \subsection{Continual Classification}
% Though construction of flashcards is completely unsupervised, it can be directly used for continual classification. Two aspects are discussed: new classes from different domains over tasks, and new instances from same domain across tasks / sessions. In both cases, we show flashcards exhibit better performance than regularization and generative replay methods, and is on-par with episodic replay.
% \vspace{-0.1cm}
\subsection{New Task Classification}
% \vspace{-0.1cm}
% \noindent \textbf{4.3.1 New Task Classification:}
Task Incremental Learning (Task-IL) assumes learning new tasks, each comprising of multiple classes ($\geq 2$). The encoder is shared with task-exclusive multihead output. Generally, classification is performed using the specific multihead, with the help of task identifier. Baseline methods built specifically in this setup perform poorly in absence of task ID (see Figure \ref{taskIL-with-without-ids}). We compare performance of flashcards for both cases, with and without task ID. We follow the same setup as provided in \citep{braininspired2020}, replacing the last dense layer with 128 neurons instead of 2000. For the first task, classifier is trained normally. AE's encoder for flashcard construction shares same architecture as classifier encoder, and is pre-initialized with its weights, and allowed to train fully along with additional decoder. Successive tasks for both networks use new samples along with the constructed flashcards, trained for 5000 iterations, optimized using Adam with lr=0.0001, and batch size of 256. Both class soft-scores and latent space regularization is applied to classifier on replay of flashcards ($\lambda=1$). In Figure \ref{taskIL-with-without-ids}, flashcards is stable with and without task ID and further beats other baseline methods in the absence of task id. Network arch. in Supplementary.

% \vspace{-0.2cm}
\subsection{New Instance Classification}
% \noindent \textbf{4.3.2 New Instance Classification:}
Single Task, New-Instance Learning (ST-NIL) classification introduced by \citep{lomonaco2017core50}, focuses on learning new instances every session while retaining the same number of classes. Performance indicates how well the model adapts to virtual concept drift across sessions. Flashcards are constructed from AE (unsupervised) per session and passed to the classifier to get predicted softmax scores as soft class-labels. It is observed that flashcards' performance is better than regularization and on-par with episodic replay, without explicitly storing exemplars in memory. (Table \ref{table:NI_classification_1}). ResNet18 is used as classifier, optimized using SGD with learning rates  of 0.001 over 20 epochs, and new sessions are introduced by adopting the same brightness and saturation from \citep{boclreplay_2020}, test set is constant across sessions.

\begin{table}[!htpb]
\small
\centering
{ 
\begin{tabular}{p{1.7cm}p{0.5cm}p{0.66cm}p{0.66cm}p{0.66cm}p{0.66cm}p{0.66cm}} \hline %|c|c|c|c|
{\textbf{Session/ Method}} & \textbf{Type} & {\textbf{1}} &
{\textbf{2}} & {\textbf{3}} &
{\textbf{4}} & {\textbf{5}}  \\\hline
% {\textbf{Arch. Type}} & {\textbf{Params}} & {\textbf{Original}} & {\textbf{AE1}} & {\textbf{AE2}} \\\hline
{Naive*} &  {-} & $67.80$ & $69.31$ & $71.37$ & $73.12$ & $73.23$\\ 
{Cumulative*} & {-} &  $67.80$& $76.13$ & $81.22$ & $81.83$ & $82.12$\\ 
% {Coreset 500} & {ER} & $67.90$ & $71.42$ & $73.84$ & $74.94$ & $75.72$\\
\hline
{EWC*} &  {Reg} & $67.80$ & $69.45$ & $72.68$ & $74.02$ & $74.31$\\
{SI*} &  {Reg} & $67.80$ & $70.48$ & $72.82$ & $74.63$ & $74.58$\\ 
{IMM*} &  {Reg} & $67.80$ & $69.69$ & $72.85$ & $74.37$ & $73.84$\\ 
{EEIL 1K*} &  {ER} & $67.80$& $71.97$ & $73.27$ & $\mathbf{74.91}$ & $74.66$\\ 
{A-GEM 1K*} & {ER} & $67.80$& $\mathbf{72.27}$ & $\mathbf{73.72}$ & $74.81$ & $\mathbf{75.15}$\\ 
\textbf{Flashcards1K} & {FR} & $67.90$ & $71.40$ & $73.34$ & $74.84$ & $74.88$\\\hline
\end{tabular}
% \vspace{-0.2cm}
\caption{ST-NIL Classification on Cifar10 using the settings described in \citep{boclreplay_2020}. * are reported from the same paper. Flashcards created from unsupervised AE performs equally well in comparison to other methods primarily built for classification. Reg=Regularization, ER=Episodic  Replay, FR=Flashcard Replay.}
\label{table:NI_classification_1}
}
\vspace{-0.3cm}
\end{table}
\vspace{-0.1cm}

\section{Conclusion}\label{conc}
% \vspace{-0.1cm}
We introduced flashcards that can capture knowledge representations of a trained AE through recursive passing of random image patterns, and showed that it can be used as alternative to original data. We further demonstrated its efficacy as a task agnostic replay mechanism for various CL scenarios, such as classification, denoising, task incremental learning, with heterogeneous datasets. Flashcard replay outperforms generative replay and regularization methods, without additional memory and training, and also performs on par with episodic replay, without storing exemplars. The intrinsic nature of flashcards allows for data abstraction which can be exploited for data privacy applications. Generalization of flashcards to other domains will foster further research.

\clearpage
\bibliographystyle{named}
\bibliography{main}

\clearpage
% \section{Acknowledgments}
% The authors would like to thank ...
\appendix

% \section{Source Code}
% We provide the source code for flashcard construction, finding the optimal iteration, application in reconstruction and task incremental classification, in the form of python notebooks. Additionally, read-only html files of the notebooks are also included for ease of access. Baselines were run directly using the repository accompanying the cited reference papers.

\section{Datasets}

The datasets used in Sequence5 are in the order - MNIST, Fashion MNIST, Cifar10, SVHN, and Omniglot. Each dataset was considered as a task, and class labels were omitted. MNIST, Fashion MNIST and Omniglot were resized to 32x32x3 (bilinear rescale and channel copied twice) to maintain the same scale as the other two datasets.

\section{Flashcards for Single-Task Scenario}
 
Algorithm \ref{algo: flashcard_general_case} provides the steps to use flashcards for single-task scenario.
 
\begin{algorithm}
\caption{Flashcard for Single-Task / General Scenario}
\label{algo: flashcard_general_case}
\begin{algorithmic}[1]
% \REQUIRE $f_{t-1}(\cdot)$: Autoencoder model for {$t-1$^{th}} task $T_{t-1}$
\REQUIRE $f_{t}(\cdot, {\boldsymbol \theta}_{t})$: AE model for task $T_{t}$
\REQUIRE $N_f$: Number of flashcards 
\REQUIRE $r$: Number of iterations through $f_{t}(\cdot, {\boldsymbol \theta}_{t})$ %over the AE model

\STATE $F_{t} \leftarrow$ \textbf{Algorithm 1} ($f_{t}(\cdot, \boldsymbol{\theta}_{t}), N_f, r$), where $F_{t}$ is the set of flashcards for task $T_t$
\STATE Train a new AE model $g_t(., {\boldsymbol \phi}_t)$ on Flashcards $F_t$,
\begin{eqnarray}
{\boldsymbol \phi}_{t}^{*} \leftarrow \argmin_{\phi_t} \frac{1}{N_f} \sum_{n=1}^{N_f} |F_{t}^{(n)} - g_t(F_{t}^{(n)}, {\boldsymbol \phi}_t)|, 
\nonumber
\end{eqnarray}
where $F_{t}^{(n)}$ is the $n^{th}$ flashcard sample
\RETURN $g_t(., {\boldsymbol \phi}_{t}^{*})$
\end{algorithmic}
\end{algorithm}

\section{Comparison between Flashcards and other Replay methods}

Table \ref{table_flashcard_comparison} provides detailed comparison between Flashcards and other replay strategies such as episodic memory and generative replay. 

\section{Benchmarking Flashcards Number of Samples}
\begin{figure}[!htbp]
	\centering
	\includegraphics[width=0.5\textwidth]{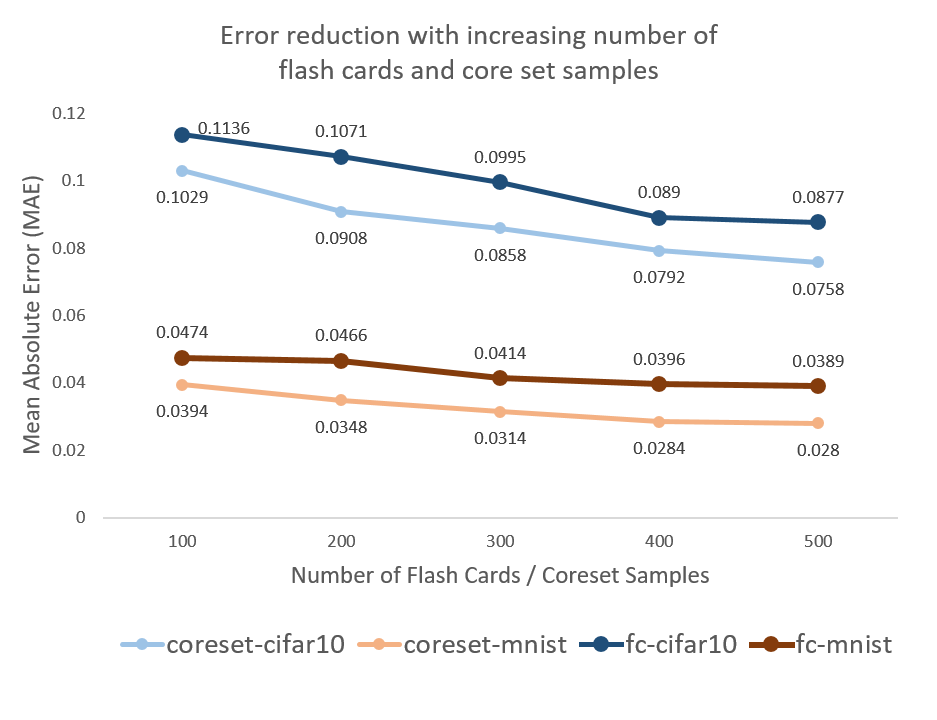}
	\caption{Increase in no. of flashcards / coreset improves performance - Quantitative error reduction in (i) Cifar10, (ii) MNIST.}
	\label{flash_card_samples_increase}
\end{figure}

Increase in flashcards number shows improvement - examples for MNIST and Cifar10, as observed in Figure \ref{flash_card_samples_increase}. Trend followed by flashcards is similar to the improvement shown with increasing coreset/exemplars.

{
\begin{table*}[]
\small
\begin{tabular}{|p{4cm}|p{4cm}| p{4cm}| p{4cm}|}
\hline
\textbf{Trait} & \textbf{Flashcards for Replay} &\textbf{Episodic Memory Replay} & \textbf{Generative (Data) Replay} \\  \hline
Examples & {\includegraphics[width=0.23\textwidth]{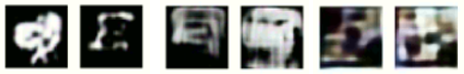}} & {\includegraphics[width=0.23\textwidth]{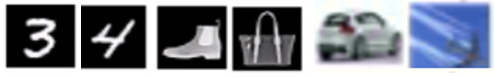}} & {\includegraphics[width=0.23\textwidth]{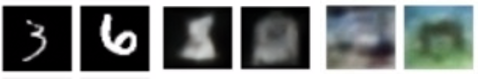}} \\   \hline
\multicolumn{4}{|c|}{\textbf{Single-Task Scenario}} \\ \hline
Alternative to original data? & Yes & No, subset of original & Yes \\   \hline
Distinction & Postprocess knowledge capture from trained network & Stores a subset of the original data in memory & Network is trained to generate samples per task \\   \hline
Visually similar to dataset? & No & Yes & Depends   on complexity \\   \hline
Performance on task & Close to original network & Very close to original network & Close to original network \\   \hline
\multicolumn{4}{|c|}{\textbf{Continual Learning Scenario}} \\  \hline
Store sample across tasks? & No & Yes, need to store per task & Depends on nw. complexity \\   \hline
Nw. used to obtain alt. data & Simple AE & N/A & VAE   / GAN etc. \\  \hline
Allocate extra memory between each task? & No, just in time creation during next task & Yes, either memory or disk & Depends on network's inbuilt support for CL \\   \hline
Constructed on the fly? & Yes, just before starting next task training & No, store original samples in memory & On the fly (if CL supported) or each task’s exclusive nw. \\   \hline
Independent of no. of tasks? & Yes, only need samples from current snapshot & No, need to store per task & Depends on network's inbuilt support for CL \\   \hline
Scaling up tasks in number and complexity & Empirical results shows strong feasibility & Straightforward;   requires linear memory storage & Scaling up in terms of no. of tasks is challenging \\   \hline
\end{tabular}
\caption{Comparison between three major replay techniques.}
\label{table_flashcard_comparison}
\end{table*}
}

\section{Using Flashcards Representation for Downstream Classification} 

\begin{table}[!h]
\small
\centering
{ 
\begin{tabular}{ |c|c|c|c|} \hline %|c|c|c|c|
{\textbf{Dataset}} & {\textbf{Original}} & {\textbf{Classifier 1}} & {\textbf{Classifier 2}} \\\hline
{MNIST} &  $0.9940$ & $0.9921$ & $0.9869$\\ 
{Fashion MNIST} &  $0.9215$ & $0.9126$ & $0.8978$\\ 
{Cifar10} &  $0.8003$ & $0.6295$ & $0.5659$\\ \hline
\end{tabular}
\caption{Building a classifier using the reconstructions from flashcard trained network. Reported accuracy averaged over 5 runs. Cifar10 accuracy is lower due to relatively higher AE recon error.}
\label{table:scratch-classification}
}
\end{table}
% We have demonstrated earlier (Figure \ref{section2_2}) that flashcards can effectively serve as alternative to the original dataset with similar performance. This experiment will further substantiate such reconstructions. 
Let AE1 be trained on original images $D_{T_1}$ and AE-Flash1 be trained using flashcards $D_{f_1}$ from AE1. Let the respective reconstructions (after training) be $\widehat{D_{T_1}}$ and $\widehat{D_{F_1}}$. Next, train two classifiers (VGG16), Classifier1 and Classifier2 using $\widehat{D_{T_1}}$ and $\widehat{D_{F_1}}$, respectively, and compare their performance on independent test set of original images.  The results tabulated in Table \ref{table:scratch-classification} shows that flashcards based networks are also capable of performing reasonably well in downstream tasks such as classification. 

\section{AutoEncoder Architecture Selection}

We train several AE architectures on Cifar10 dataset to compare the performance of flashcards for reconstruction. Table \ref{table:architecture-selection} provides the details about various model architectures and the corresponding test Mean Absolute Error (MAE) on training using the original dataset (Original MAE) and using Flashcards generated from the trained AE (Flashcards MAE), respectively. 

We choose the architecture \emph{Blk\_4\_fil\_64} for all our experiments for reconstruction and denoising tasks. \emph{Blk\_4\_fil\_64} architecture obtains $0.0512$ Original MAE and $0.0570$ Flashcards MAE. \emph{Blk\_3\_fil\_64} and \emph{Blk\_2\_fil\_32} achieve better Original MAE/Flashcards MAE than \emph{Blk\_4\_fil\_64}, but both these architectures use higher latent space size ($1024$ and $2048$). The improvement in reconstruction MAE may be attributed to such high latent space dimensions which might not encode useful information and just act as a copy function. Hence, we choose the \emph{Blk\_4\_fil\_64} architecture.       

\begin{table*}[htpb]
\small
\centering
{ 
\begin{tabular}{ |c|c|c|c|c|c|c|} \hline
% \multirow{2}{*}{\textbf{Arch. Type}} & \multirow{2}{*}{ Params} & \multirow{2}{*}{\textbf{Original}} & \multirow{2}{*}{ \begin{tabular}{@{}c@{}} \textbf{AE1} \\ \textbf{} \end{tabular}} & \multirow{2}{*}{ \begin{tabular}{@{}c@{}} \textbf{Flashcards} \\ \textbf{from AE2} \end{tabular}} \\ {} & {} & {} & {} & {} \\\hline
{\textbf{Arch. Type}} & {\textbf{Model Params.}} & {\textbf{Latent Space}} & {\textbf{Num. Blocks}} & {\textbf{Num. Filters}} & {\textbf{Original MAE}} &  {\textbf{Flashcards MAE}} \\\hline
{Blk\_4\_fil\_16} & $24, 083$ & $64$ (48x reduction) & $4$ & $16$ & $0.0787 \pm 0.0002$ & $0.0963 \pm 0.0004$ \\ \hline 
{Blk\_4\_fil\_32} & $94, 243$ & $128$ (24x reduction) & $4$ & $32$ & $0.0640 \pm 0.0002$ & $0.0725 \pm 0.0009$ \\ \hline
{\textbf{Blk\_4\_fil\_64}} & $372, 803$ & $\textbf{256}$ \textbf{(12x reduction)} & $4$ & $64$ & $0.0512 \pm 0.0004$ & $0.0570 \pm 0.0006$ \\ \hline
{Blk\_4\_fil\_128} & $1, 482, 883$ & $512$ (6x reduction) & $4$ & $128$ & $0.2062 \pm 0.0000$ & $0.2445 \pm 0.0476$ \\\hline
{Blk\_3\_fil\_64} & $298, 947$ & $1024$ (3x reduction) & $3$ & $64$ & $0.0437 \pm 0.0003$ & $0.0599 \pm 0.0067$ \\\hline
{Blk\_2\_fil\_32} & $57, 251$ & $2048$ (1.5x reduction) & $2$ & $32$ & $0.0358 \pm 0.0008$ & $0.0389 \pm 0.0015$ \\\hline
\end{tabular}
\caption{Architecture selection for AutoEncoder (AE). We train several AE architectures on Cifar10 dataset in order to compare the performance of flashcards for reconstruction. Various details about the architecture such as Model Params. (Number of trainable weights and biases), Latent Space (Size of latent space/bottleneck layer and its reduction rate versus image space), Num. Blocks (Number of convolution + pooling blocks in Encoder), and Num. Filters (Number of filters in convolution layers) are also provided. Original MAE is the Cifar10 test MAE on AE trained using Cifar10 train dataset. Flashcards MAE is the Cifar10 test MAE on AE trained using the flashcards obtained from given AE. The reported standard deviation for the scores are obtained over 5 experimental runs.}
\label{table:architecture-selection}
}
\end{table*}

\section{Redefining Metrics for Reconstruction}
 We measure individual task MAE before and after observing the data for the given task as well as the ability of transferring knowledge from one task to another, by measuring BWT (Backward Transfer) and FWT (Forward Transfer). We follow the similar definition for avg mean absolute error (MAE), BWT, and FWT as described in \citep{GEM}. It must be noted that \citep{GEM} describes these definitions in a supervised setting. As we explore an unsupervised representation, we redefine these metrics for unsupervised settings. Consider we have the test sets for each of the $T$ tasks. We evaluate the model obtained after training task $T_t$ on all $T$ tasks. This gives us the matrix $M \in R^{{T \times T}}$, where $M_{i, j}$ represents the test MAE for model on task $T_j$ after observing the data of task $T_i$. Let $r$ be vector of test MAEs for each task obtained using random weight initializations. Then, we define following three metrics,

\begin{equation}
    \label{eqn: avg_mae}
    Avg MAE = \frac{1}{T}\sum\limits_{i=1}^T M_{T, i} 
\end{equation}
\begin{equation}
    \label{eqn: bwt}
    BWT = \frac{1}{T - 1}\sum\limits_{i=1}^{T-1} (M_{i, i} - M_{T, i}) 
\end{equation}
\begin{equation}
    \label{eqn: fwt}
    FWT = \frac{1}{T-1}\sum\limits_{i=2}^{T} (r_{i} - M_{i-1, i}) 
\end{equation}

Lower value for Avg MAE and higher values for BWT and FWT are better.

\section{Sequence5 Continual Reconstruction}
\begin{figure*}[htpb]
	\centering
	\includegraphics[width=0.9\textwidth]{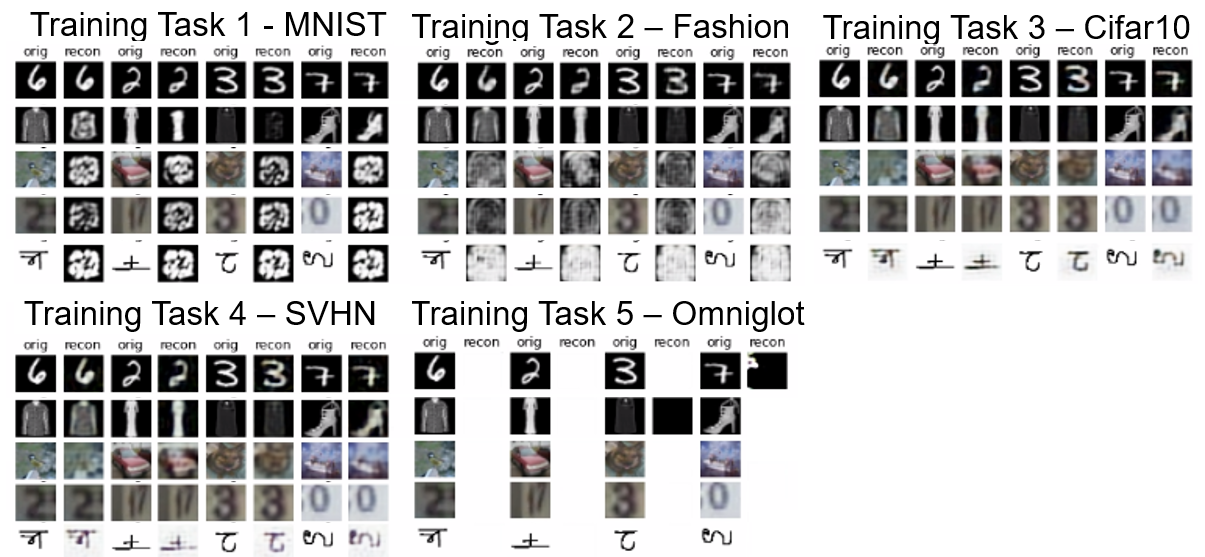}
	\caption{Continual Reconstruction on Naive / Sequential Fine Tuning (SFT).}
	\label{cl_reconstruction_sft}
\end{figure*}

\begin{figure*}[htpb]
	\centering
	\includegraphics[width=0.9\textwidth]{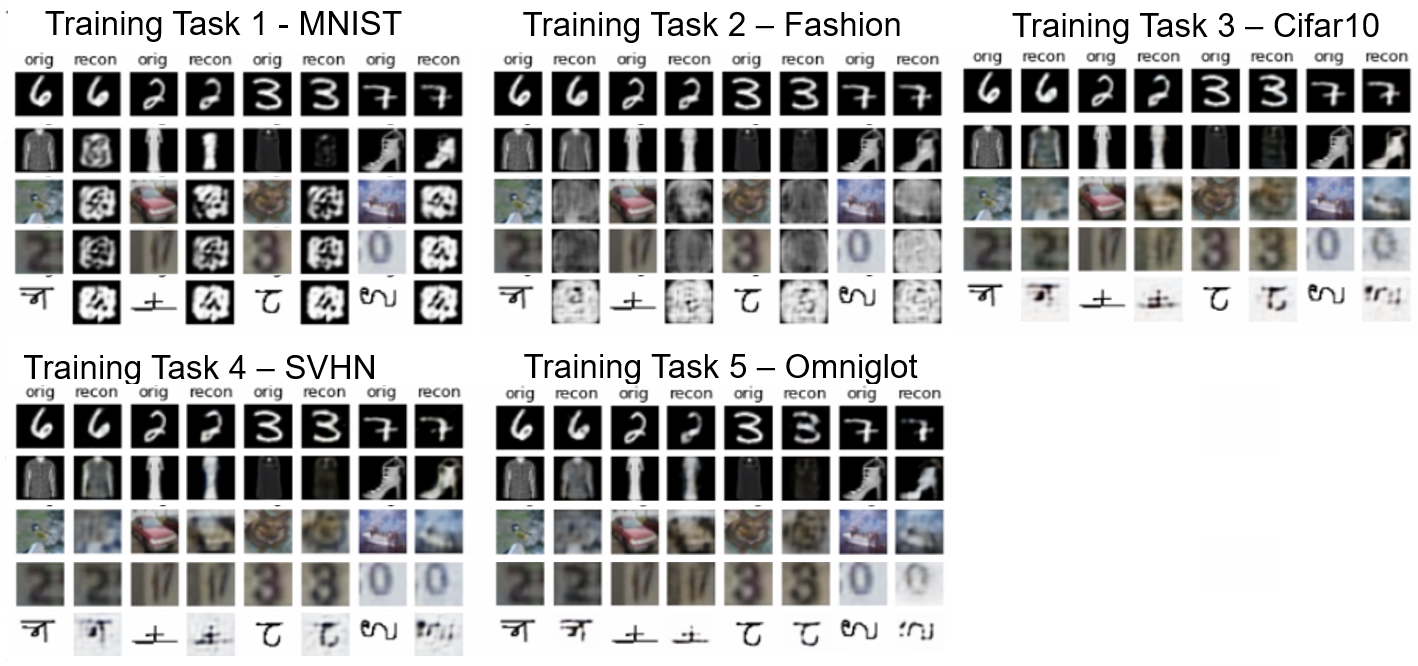}
	\caption{Continual Reconstruction using episodic memory - coreset 500.}
	\label{cl_reconstruction_coreset}
\end{figure*}

\begin{figure*}[htpb]
	\centering
	\includegraphics[width=1.0\textwidth]{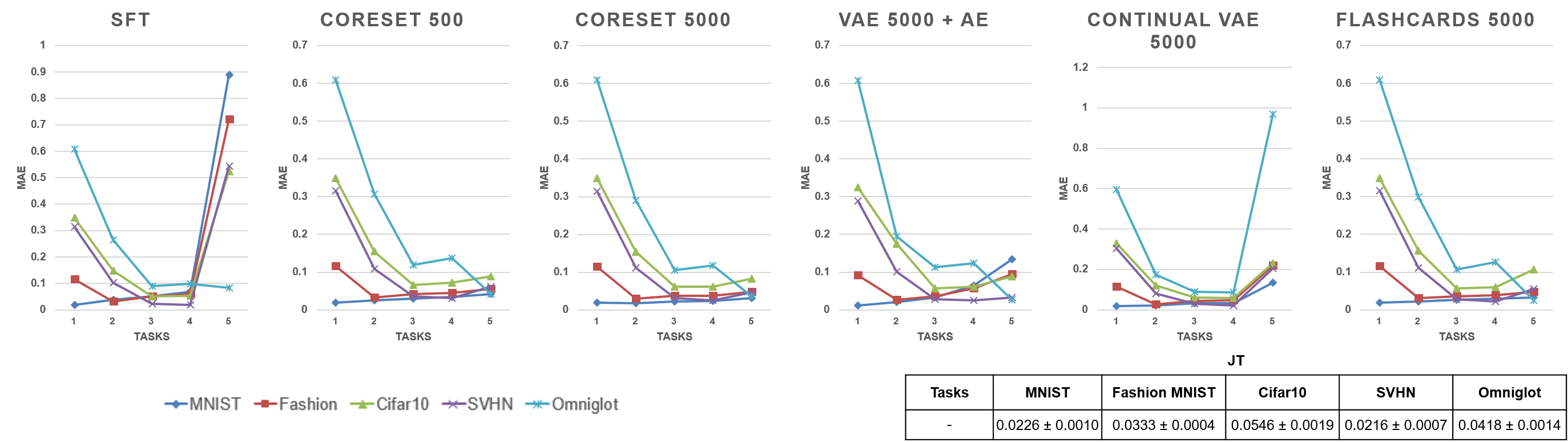}
	\caption{Continual Learning for Reconstruction. Individual graphs for different methods show the variation of test Mean Absolute Error (MAE) on current task dataset after observing the data for sequence of tasks. The table shows the test MAE for Joint Training (JT) method. The reported values in graph and table obtained over 5 experimental runs. The standard deviation is quite small and is not displayed on the graphs to avoid clutter.}
	\label{results_recon_graph}
\end{figure*}

We provide visuals for each task in Sequence5 showing how each method handles forgetting. Figure \ref{cl_reconstruction_sft} shows Sequential Fine Tuning (SFT) is the naive approach and suffers the most. It can be observed that reconstructions are empty, this is because of the network parameters at the start of task 5, which prevents it to learn the current Omniglot task itself. Figure \ref{cl_reconstruction_coreset} shows the effect of replay with 500 real samples (coreset). 500 samples were chosen as their memory matches the AE network parameters of 1.5MB. From the experimental results, it is observed that 500 samples are not sufficient to beat flashcards. Figure \ref{results_recon_graph} has individual graphs for different methods show the variation of test Mean Absolute Error (MAE) on current task dataset after observing the data for sequence of tasks.

\begin{figure*}[htpb]
	\centering
	\includegraphics[width=0.9\textwidth]{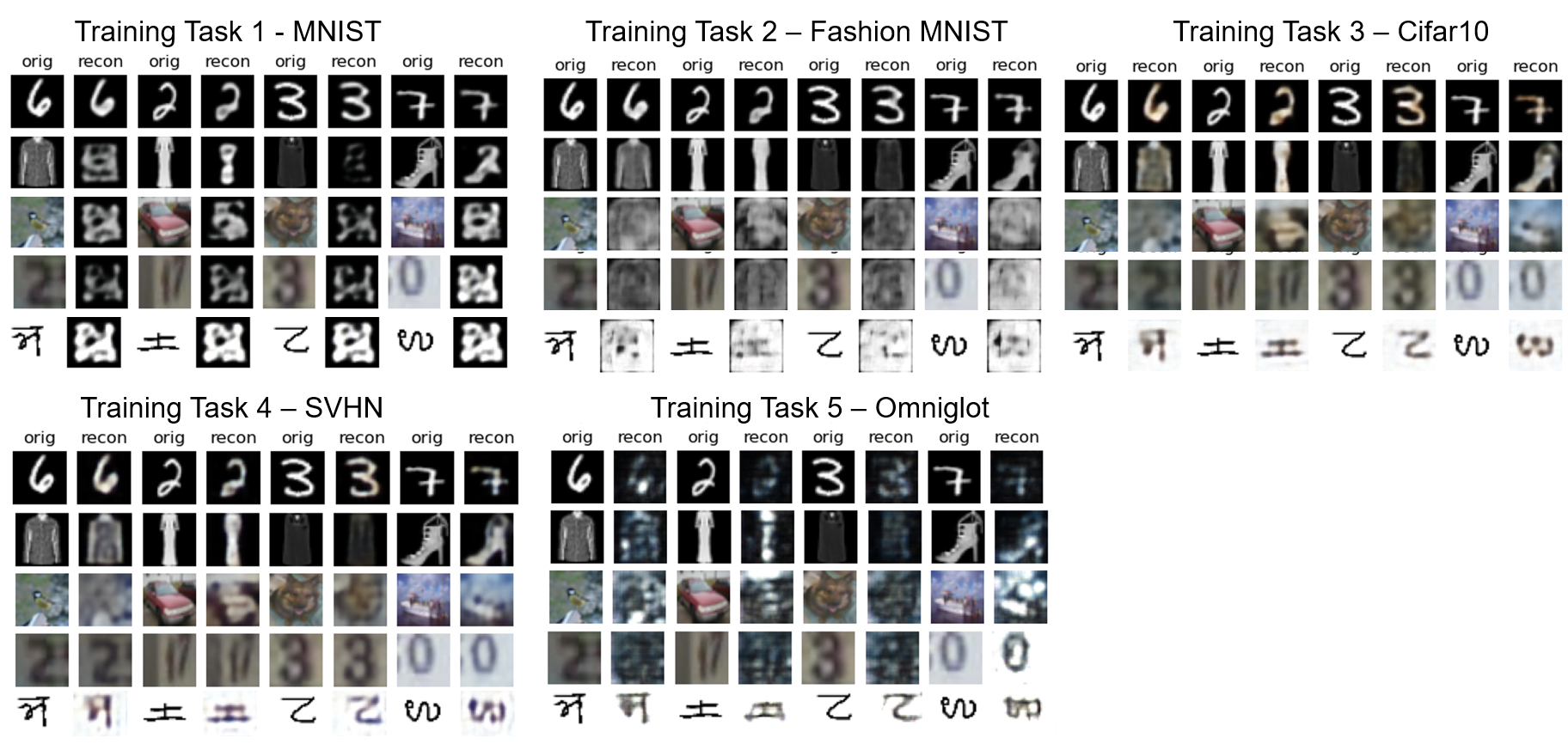}
	\caption{Continual Reconstruction using VAE trained exlusively for Continual Learning.}
	\label{cl_reconstruction_clvae}
\end{figure*}

\begin{figure*}[htpb]
	\centering
	\includegraphics[width=0.9\textwidth]{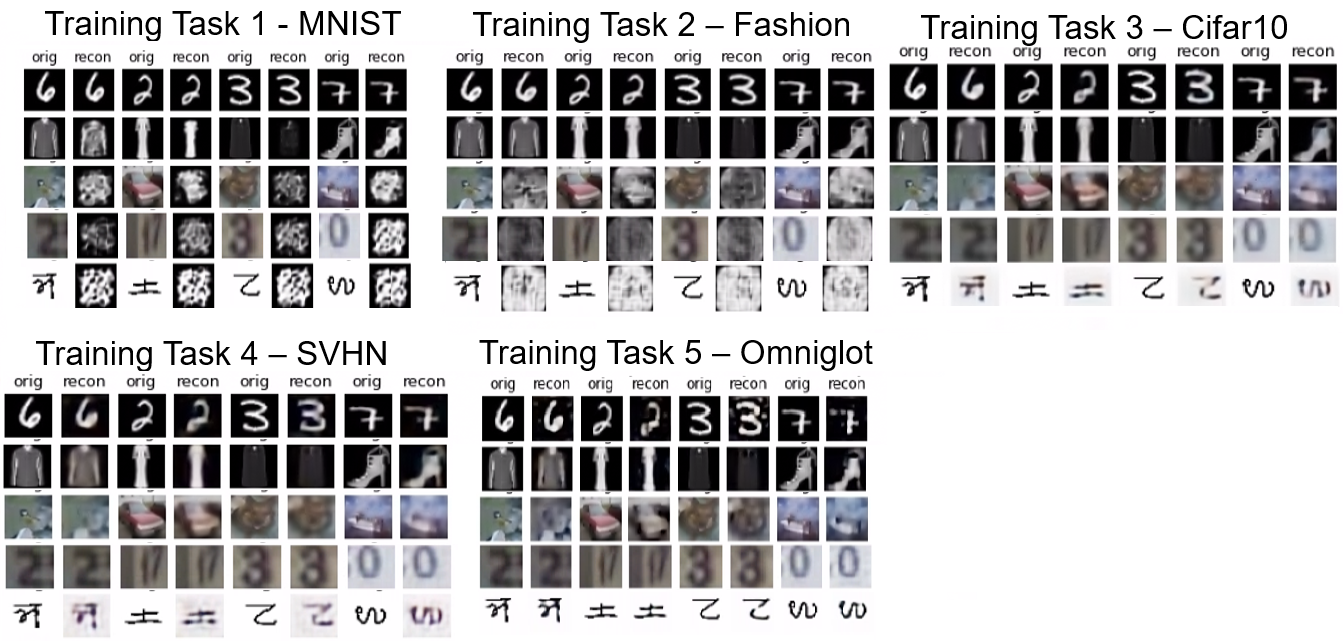}
	\caption{Continual Reconstruction using AE + VAE as generative replay.}
	\label{cl_reconstruction_ae_vae}
\end{figure*}

\begin{figure*}[htpb]
	\centering
	\includegraphics[width=0.9\textwidth]{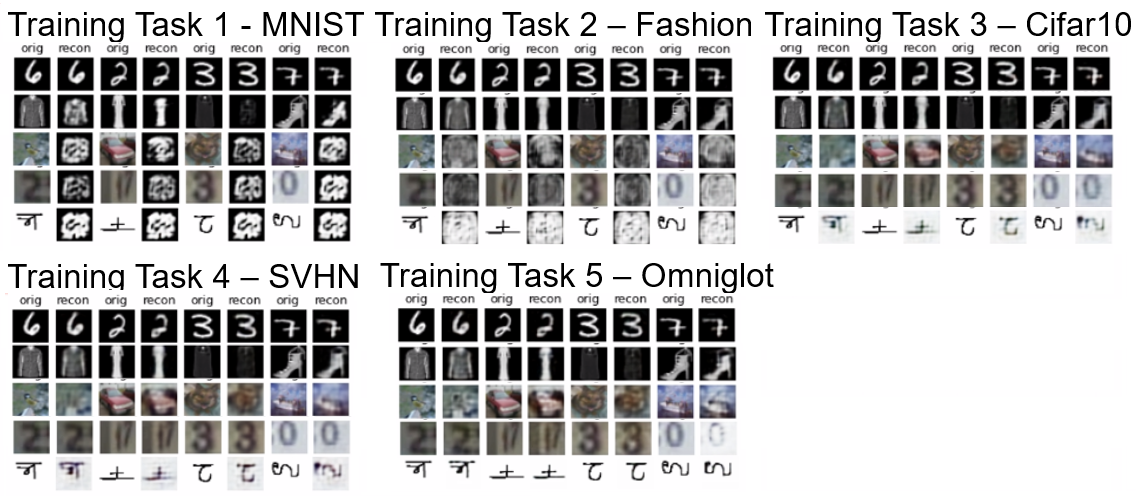}
	\caption{Continual Reconstruction using Flashcards.}
	\label{cl_reconstruction_flashcards5000}
\end{figure*}

Figure \ref{cl_reconstruction_clvae} is based on VAE trained in CL fashion, maintaining the same mean and std.dev. across tasks. It is not sufficient to mitigate forgetting. 
Figure \ref{cl_reconstruction_ae_vae} uses AE for reconstruction supplemented by an external VAE for generative replay. Though results are competitive with Flashcards, there is still forgetting in the previous tasks - MNIST and Fashion MNIST.
Figure \ref{cl_reconstruction_flashcards5000} presents results when using Flashcards, where the past and current task samples are remembered well. 

\section{Sequence3 Continual Reconstruction}
We also permuted the order by taking MNIST, Fashion MNIST and Cifar10, to further substantiate the effect of flashcards. Tables 5, 6, and 7
% \ref{table:seq3-mnist}, \ref{table:seq3-fashion} and \ref{table:seq3-cifar10} 
compare the mitigation of forgetting with flashcards.

\begin{table*}[htpb]
\centering{
\begin{tabular}{|c |c |c |c |c |c |}
\hline
\textbf{Method} & \textbf{Task} & \textbf{MNIST} & \textbf{Fashion MNIST} & \textbf{Cifar10} & \textbf{Avg MAE} \\ \hline
Joint Training & - & 0.0141 & 0.0256 & 0.0629 & 0.0342 \\ \hline
& 1 & 0.0190 & - & - & 0.0190 \\
{Coreset Sampling 5000} & 2 & 0.0245 & 0.0388 & - & 0.0316 \\ 
  & 3 & 0.0249 & 0.0395 & 0.0666 & 0.0436 \\ \hline
&  1 & 0.0190 & - & - & 0.0190 \\
{Lower Bound} &  2 & 0.0268 & 0.0268 & - & 0.0268 \\ 
& 3 & 0.0467 & 0.0469 & 0.0512 & 0.0482 \\ \hline
& 1 & 0.0190 & - & - & 0.0190 \\
{Flashcards 5000} & 2 & 0.0243 & 0.0310 & - & 0.0276 \\ 
 & 3 & 0.0282 & 0.0366 & 0.0579 & \textbf{0.0409} \\ \hline
\end{tabular}
\label{table:seq3-mnist}
\caption{Sequence of 3 - Order: MNIST, Fashion MNIST, Cifar10 as 3 tasks. Tasks are added incrementally, and MAE is computed on each dataset after current task is completed.}
}\end{table*}

\begin{table*}[htpb]
\centering
\begin{tabular}{|c |c |c |c |c |c |}
\hline
\textbf{Method} & \textbf{Task} & \textbf{Fashion MNIST} & \textbf{Cifar10} & \textbf{MNIST} & \textbf{Avg MAE} \\ \hline
Joint Training & - &  0.0141 & 0.0256 & 0.0629 & 0.0342\\ \hline
& 1 & 0.0324 & - & - & 0.0324 \\
{Coreset Sampling 5000} & 2 & 0.0344 & 0.0589 & - & 0.0466 \\ 
  & 3 & 0.0386 & 0.0661 & 0.0200 & 0.0415 \\ \hline
&  1 & 0.0324 & - & - & 0.0324 \\
{Lower Bound} &  2 & 0.0548 & 0.0564 & - & 0.0556 \\ 
 & 3 & 0.0816 & 0.2996 & 0.0140 & 0.1317 \\ \hline
& 1 & 0.0324 & - & - & 0.0324 \\
{Flashcards 5000} & 2 & 0.0336 & 0.0520 & - & 0.0428 \\ 
 & 3 & 0.0352 & 0.0637 & 0.0156 & \textbf{0.0381} \\ \hline
\end{tabular}
\label{table:seq3-fashion}
\caption{Sequence of 3 - Order: Fashion MNIST, Cifar10, MNIST as 3 tasks. }
\end{table*}

\begin{table*}[htpb]
\centering{
\begin{tabular}{|c |c |c |c |c |c |}
\hline
\textbf{Method} & \textbf{Task} & \textbf{Cifar10} & \textbf{MNIST} & \textbf{Fashion MNIST} & \textbf{Avg MAE} \\ \hline
Joint Training & - &  0.0141 & 0.0256 & 0.0629 & 0.0342\\ \hline
& 1 & 0.0515 & - & - & 0.0515 \\
{Coreset Sampling 5000} & 2 & 0.0639 & 0.0220 & - & 0.0429 \\ 
  & 3 & 0.0654 & 0.0229 & 0.0336 & \textbf{0.0406} \\ \hline
&  1 & 0.0515 & - & - & 0.0515 \\
{Lower Bound} &  2 & 0.2602 & 0.0142 & - & 0.1372 \\ 
 & 3 & 0.1233 & 0.0465 & 0.0371 & 0.0689 \\ \hline
& 1 & 0.0515 & - & - & 0.0515 \\
{Flashcards 5000} & 2 & 0.0625 & 0.0181 & - & 0.0403 \\ 
 & 3 & 0.0664 & 0.0261 & 0.0308 & 0.0411 \\ \hline
\end{tabular}
}
\label{table:seq3-cifar10}
\caption{Sequence of 3 - Order: Cifar10, MNIST, Fashion MNIST as 3 tasks. }
\end{table*}

\section{Sequence5 Continual Denoising - Adjusting the weight of noise factor}
\begin{figure*}[htpb]
	\centering
	\includegraphics[width=0.9\textwidth]{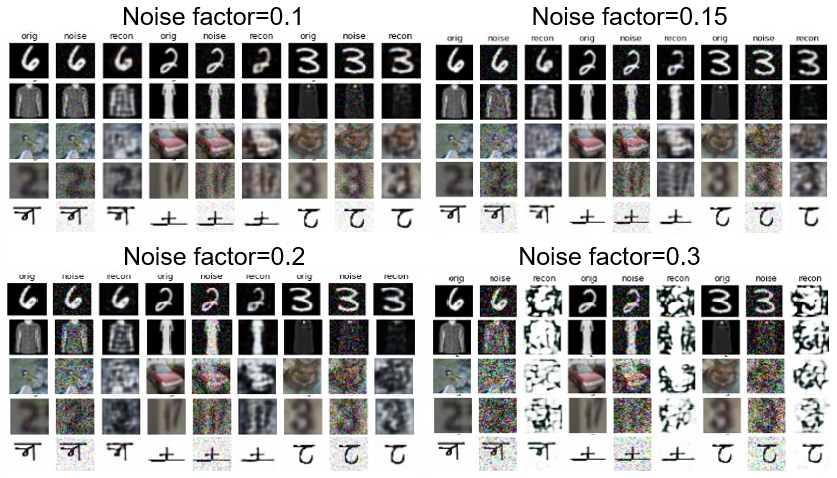}
	\caption{Continual Denoising scenario. Shown in figure is the effect of noise applied and the reconstruction of Sequence5 using flashcards.}
	\label{noise_factors_denoising}
\end{figure*}

We increased the noise factor steadily to check for the value where reconstruction fails completely. Figure \ref{noise_factors_denoising} shows the impact of reconstruction using flashcards for different noise level settings. As more noise is added, it becomes visually difficult to make out the underlying image. At factor of 0.3, it is observed the network is trying to retain partial outer boundary but has forgotten the denoising ability when seeing the last task Omniglot.

\begin{table*}[!htpb]
\small
\centering
{ 
\begin{tabular}{ |c|c|c|c|c|c|} \hline %|c|c|c|c|
\textbf{Method/Sess} & \textbf{1} &
\textbf{2} &\textbf{3} &
\textbf{4}&
\textbf{5} \\ {} & {} & {} & {} & {} & {} \\\hline
% {\textbf{Arch. Type}} & {\textbf{Params}} & {\textbf{Original}} & {\textbf{AE1}} & {\textbf{AE2}} \\\hline
\multicolumn{6}{c}{\textbf{Cifar10}} \\ \hline
% {Cifar10}  \\ \hline
{Naive} &  $67.90 \pm 0.86$ & $67.35 \pm 0.81$ & $67.66 \pm 0.23$ & $61.18 \pm 0.96$ & $63.05 \pm 0.86$\\ 
{Cumulative} &  $67.90 \pm 0.86$ & $69.74 \pm 0.52$ & $72.00 \pm 0.28$ & $72.85 \pm 0.37$ & $73.54 \pm 0.22$\\
{Coreset500} &  $67.90 \pm 0.86$ & $67.50 \pm 0.24$ & $68.29 \pm 0.18$ & $61.97 \pm 0.82$ & $63.85 \pm 0.07$\\
{Coreset5000} &  $67.90 \pm 0.86$ & $\textbf{68.47} \pm \textbf{0.51}$ & $\textbf{68.38} \pm \textbf{0.79}$ & $64.12 \pm 0.46$ & $65.18 \pm 0.93$\\ \hline
\textbf{{Flashcards}} &  $67.90 \pm 0.86$ & $68.05 \pm 0.95$ & $68.14 \pm 1.27$ & $\textbf{65.64} \pm \textbf{1.31}$ & $\textbf{65.83} \pm \textbf{1.07}$\\ \hline
\multicolumn{6}{c}{\textbf{Cifar100}} \\ \hline
{Naive} &  $37.67 \pm 0.31$& $34.03 \pm 0.53$ & $35.10 \pm 0.23$ & $25.35 \pm 0.69$ & $28.74 \pm 0.33$\\ 
{Cumulative} &  $37.67 \pm 0.31$& $40.63 \pm 0.15$ & $41.02 \pm 0.18$ & $41.24 \pm 0.20$ & $41.55 \pm 0.11$\\
{Coreset500} &  $37.67 \pm 0.31$& $34.03 \pm 0.51$ & $34.95 \pm 0.24$ & $25.79 \pm 1.55$ & $29.09 \pm 0.21$\\
{Coreset5000} &  $37.67 \pm 0.31$& $34.97 \pm 0.08$ & $35.02 \pm 0.48$ & $29.97 \pm 0.07$ & $30.63 \pm 0.44$\\ \hline
\textbf{Flashcards} &  $37.67 \pm 0.31$ & $\textbf{36.53} \pm \textbf{0.22}$ & $\textbf{36.34} \pm \textbf{0.50}$ & $\textbf{33.47} \pm \textbf{1.07}$ & $\textbf{32.60} \pm \textbf{0.67}$\\\hline
\end{tabular}
\caption{ST-NIL Classification on Cifar10 and Cifar100 with mean and std.dev. over 5 runs.}
\label{table:NI_classification_2}
}
\end{table*}

\section{New Task Incremental Learning}
This section provides architectural and training details for the Task-IL classification. Architecture of classifier is based on specifications from \citep{braininspired2020}, and has five convolutional layers containing 16, 32, 64, 128 and 254 channels. Each layer used a 3×3 kernel, a padding of 1, and there was a stride of 1 in the first layer (i.e., no downsampling) and a stride of 2 in the other layers (i.e., image-size was halved in each of those layers). All convolutional layers used batch-norm followed by a ReLU non-linearity. The last conv layer is flattened to give 1024 dimensions, followed by a dense layer of 2000 activation followed by another dense layer of 128, forming the latent space. The output layer nodes were added based on the task and class information. Architecture for AE - encoder of classifier is shared with weights of trained classifier's encoder and pre-initialized, but allowed to train further. Decoder is attached after latent space, and AE is trained by passing current task samples along with flashcards from AE upto current task.

Training details: For all tasks, mini-batch size of 256 and Adam optimizer with lr=0.0001 is used. Both models are trained for 5000 iterations. For 1st task, train classifier normally using crossentropy loss. Copy weights of classifier's encoder to AE encoder. Train AE using mean square error (MSE) loss. From next task onwards, train classifier for new task along with constructed flashcards from AE of previous task. Output of flashcards is obtained by passing them to the classifier and obtaining soft labels, before the start of next task. Additionally, latent space activation can also be used to regularize the network further. After classifier is trained, copy weights of encoder to AE encoder and train AE with both current task and flashcards. Repeat process until end of all tasks.

\section{Single Task - New Instance Classification Forgetting Setting}
We notice that the experimental setting mentioned in \citep{boclreplay_2020} does not reflect the forgetting scenario. 
% Even visually, there does not appear to be much change across sessions Figure \ref{ni_cls_figure}. 
Therefore, we introduce a more challenging setting across sessions: brightness\_jitter = [0, -0.1, 0.1, -0.2, 0.2] and saturation\_jitter = [0, -0.1, 0.1, -0.2, 0.2]. We employ the same learning parameters as described in their work, using ResNet18 as the classifier and optimizing using SGD with learning rates of 0.001 over 20 epochs. Results for Cifar10 and Cifar100 in Table \ref{table:NI_classification_2} show replay using flashcards is at-par with episodic coreset replay.

% \clearpage

% \bibliographystyle{named}
% \bibliography{main}

% \clearpage

\clearpage

\end{document}